%% file: main.tex
\documentclass[10pt,twocolumn,letterpaper]{article}

\usepackage[pagenumbers]{cvpr} 

\input{preamble}

\definecolor{cvprblue}{rgb}{0.21,0.49,0.74}
\usepackage[pagebackref,breaklinks,colorlinks,allcolors=cvprblue]{hyperref}

\usepackage{siunitx}
\usepackage{bm}
\usepackage{graphicx}
\usepackage{amsmath}
\usepackage{amssymb}
\usepackage{booktabs}
\usepackage{comment}
\newtheorem{remark}{Remark}

\graphicspath{{figures_overleaf/}}

\input{math_macros}
\input{ALL_FIGURES}
\input{ALL_TABLES}

\def\paperID{7799} 
\def\confName{CVPR}
\def\confYear{2026}

\title{Hearing the Room Through the Shape of the Drum: Modal-Guided Sound Recovery from Multi-Point Surface Vibrations\vspace*{-5mm}}

\author{Shai Bagon \qquad Matan Kichler \qquad Mark Sheinin\\
Weizmann Institute of Science, Israel\\
{\tt\small shai.bagon@weizmann.ac.il, matankic@gmail.com,  mark.sheinin@weizmann.ac.il}\\
{\small
\url{https://shaibagon.github.io/hearing_the_shape_of_the_drum}}
}

\begin{document}
\figTeaser  
\maketitle
\input{sec/0_abstract}    
\input{sec/1_intro}

\input{sec/2_background}

\input{sec/3_recover_sound}

\input{sec/4_robust_mode_estimation}
\input{sec/4p5_calibrated}

\input{sec/5_implement_details}
\input{sec/6_experimental_evaluation}
\vspace{-12pt}
\input{sec/7_disscusion}
\input{sec/8_conclusion}

\paragraph{Acknowledgments:}
We thank Prof.~Sharon Gannot, Aviya Shmuel, and Maya Moriya. 
This project received funding from the NSF-BSF grant \#2024808, the MBZUAI-WIS Joint Program for AI Research,
and the Knell Family Institute for Artificial Intelligence.

\newpage
{
    \small
    \bibliographystyle{ieeenat_fullname}
    \bibliography{main}
}
\newpage
\appendix
\input{sec/X_suppl}

\end{document}

%% file: preamble.tex
\newcommand{\afterfigure}{\vspace*{-8pt}}

\usepackage[percent]{overpic}

\usepackage{xspace}
\usepackage{ifthen}
\usepackage[normalem]{ulem}

\newcommand{\SB}[2][]{%
    \ifthenelse{ \equal{#1}{} }
        {\textcolor{orange}{#2\xspace{}}}
        {\textcolor{orange}{\sout{#1} #2\xspace{}}}
}
\newcommand{\MT}[2][]{%
    \ifthenelse{ \equal{#1}{} }
        {\textcolor{blue}{#2\xspace{}}}
        {\textcolor{blue}{\sout{#1} #2\xspace{}}}
}

\definecolor{darkred}{rgb}{0.75, 0.1, 0.5}
\newcommand{\CR}[2][]{%
    \ifthenelse{ \equal{#1}{} }
        {\textcolor{darkred}{#2\xspace{}}}
        {\textcolor{darkred}{\sout{#1} {#2}\xspace{}}}
}

\DeclareMathOperator*{\argmin}{arg\,min}

\def\eg{\emph{e.g.}} 

\def\ie{\emph{i.e.}} 

\def\etal{\emph{et al.}}

\def\capfont{\normalfont\small}

%% file: math_macros.tex
\newcommand{\modeShapeGrad}{\nabla\phi_k(\mathbf{x}_n)}  
\newcommand{\modeImpulseRes}{g_k(t)} 

%% file: ALL_FIGURES.tex
\newcommand{\figTeaser}{
\twocolumn[{
\renewcommand\twocolumn[1][]{##1}
\maketitle
\centering
\vspace*{-8mm}
\includegraphics[width=\linewidth]{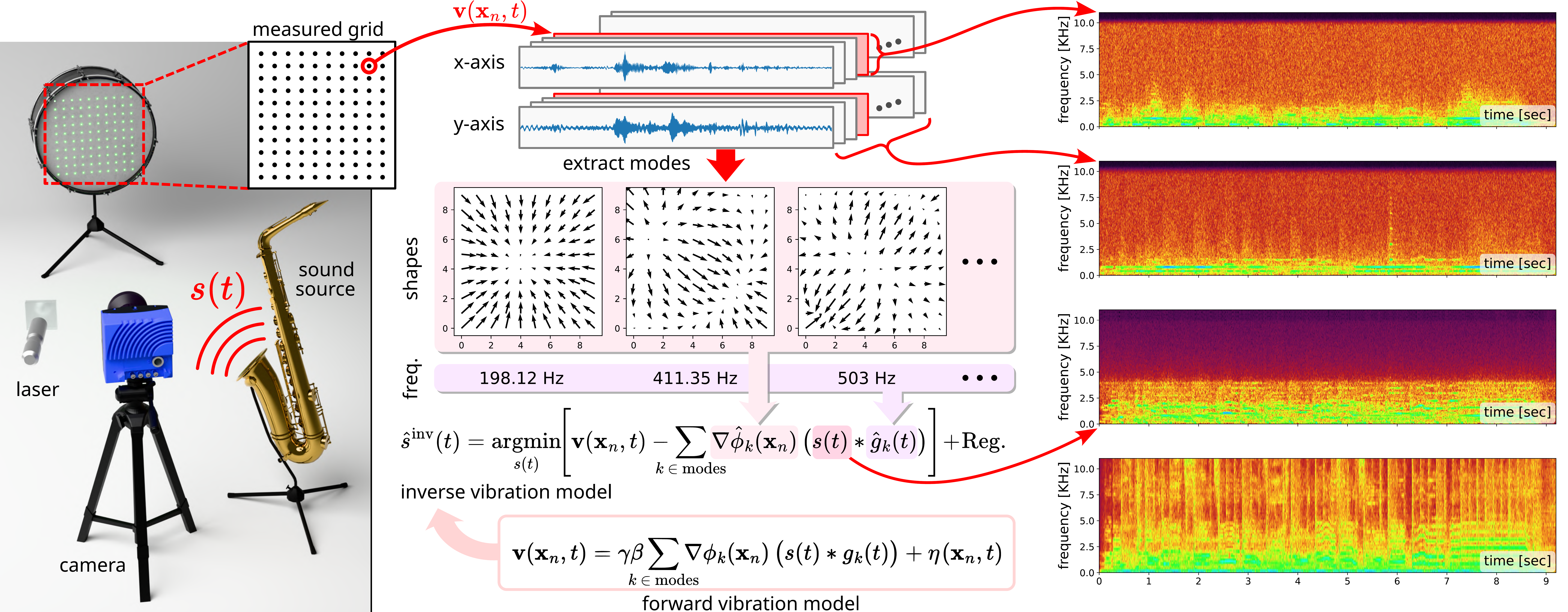}
\put(-115,146.5){\capfont single-axis signal \cite{Sheinin:2022:Vibration}}
\put(-114,99.5){\capfont average of all points}
\put(-94,51.5){\capfont our \( \hat s^{\rm inv}(t) \)}
\put(-108,2){\capfont source signal \(s(t)\)}
\vspace*{-0.1cm}
\captionof{figure}{We introduce a novel approach for sound recovery from multi-point, speckle-based vibration measurements. Our system captures a grid of speckle-based vibration signals across an object's surface.
We derive a novel vibration forward model connecting the multi-point two-axis measurements to the underlying scene sound source. The model relies on extracting the object's vibrational modes from the data. Then, we invert the model to estimate the scene sound source via optimization, yielding superior sound recoveries.}
\label{fig:teaser}
\vspace*{0.26cm}
}]
}

\newcommand{\figSignals}{
\begin{figure}[t]
    \centering
    \includegraphics[width=0.9\columnwidth]{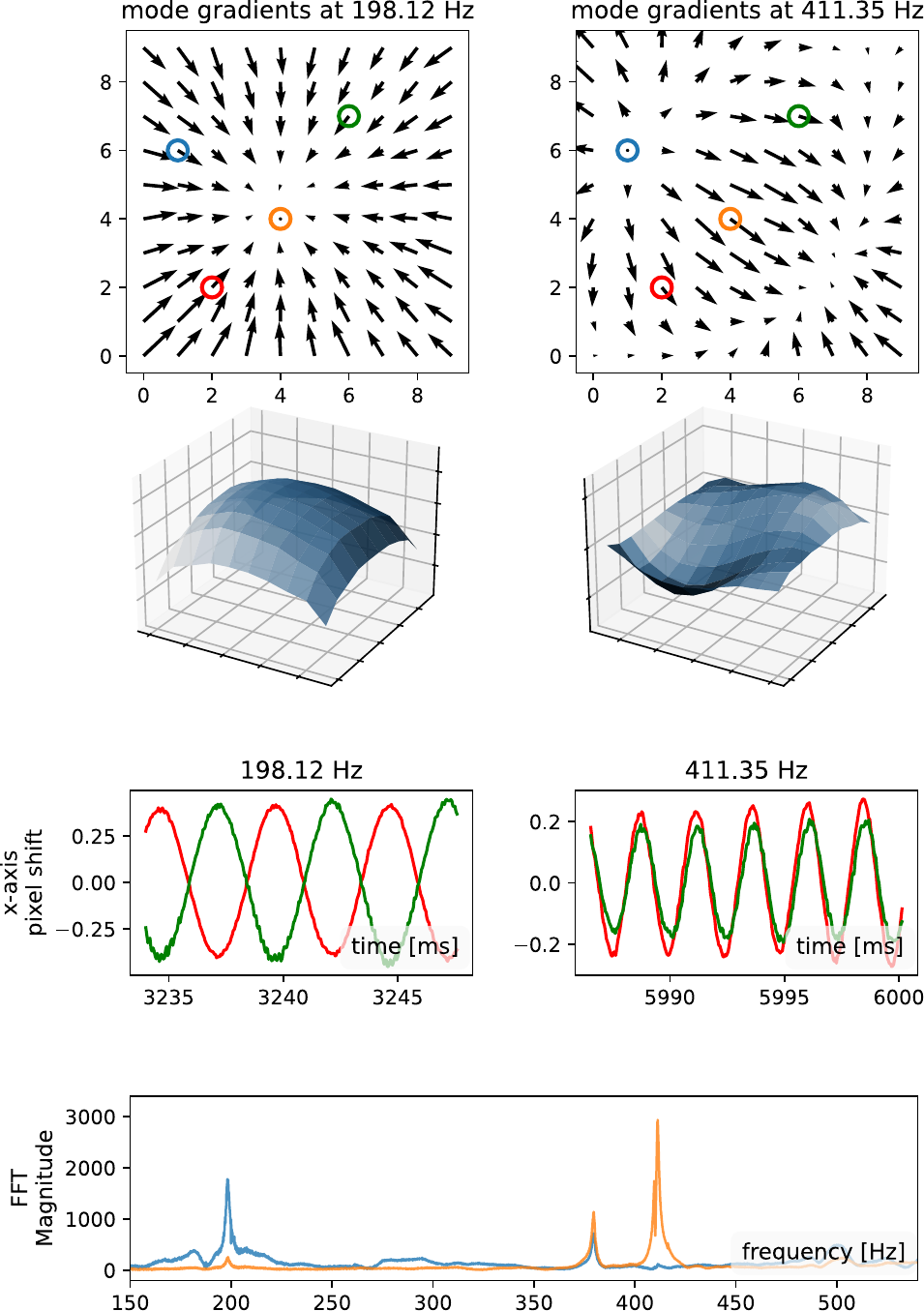}
    \put(-205,135){\capfont \textbf{(a)} mode gradients (measured) and shapes (computed)}
    \put(-205,61){\capfont \textbf{(b)} x-axis speckle shifts for both modes at two points}
    \put(-205,-9){\capfont \textbf{(c)} x-axis FFT magnitudes for blue and orange points }
    \caption{Frequency-dependent coupling of speckle shifts across surface points.
    We measure drum membrane vibrations in response to a logarithmic chirp using a $10\!\times\!10$ speckle grid. Different audio frequencies produce distinct vibration patterns.
    \textbf{(a)}~Two-axis speckle shifts at two resonant frequencies (top) and their corresponding integrated mode shapes (bottom).
    \textbf{(b)}~Speckle shifts are frequency-coupled: the red and green points oscillate out of phase at \SI{198}{Hz} but in phase at \SI{411}{Hz}, so averaging them suppresses \SI{198}{Hz} components.
    \textbf{(c)}~Points near mode-shape peaks or valleys (where gradients vanish) show weak signal energy at that mode, e.g., the blue point lacks spectral content near \SI{411}{Hz}.
    }
    \afterfigure
    \label{fig:signals}
\end{figure}
}

\newcommand{\figFindModes}{
\begin{figure}
\centering
\begin{overpic} 
    [width=\linewidth]{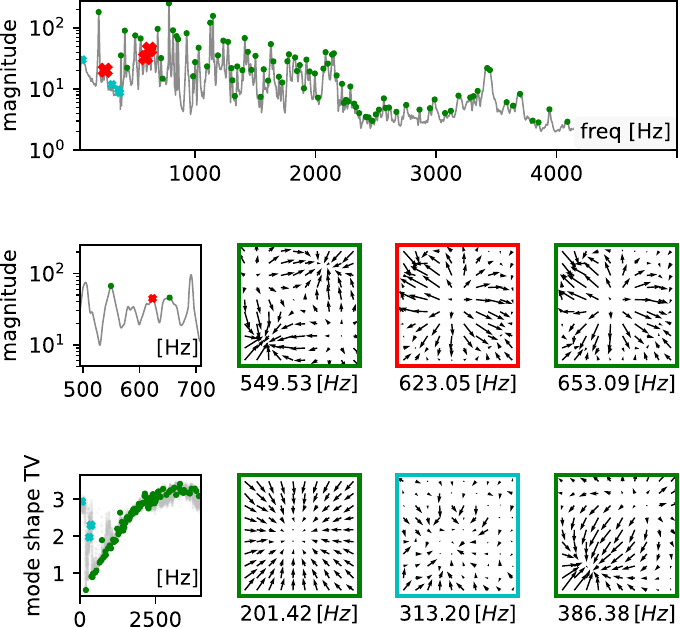}
    \put(50, 87){$\sigma(\omega)$}
    \put(6, 61){\capfont \textbf{(a)} candidate mode frequencies from smoothed spectrum}
    \put(6, 29){\capfont \textbf{(b)} removal of redundant modes with correlated shapes}
    \put(6, -5){\capfont \textbf{(c)} elimination of physically inconsistent high-TV modes}
\end{overpic}
\vspace*{0mm}
\caption{Robust mode estimation.
(a)~Initial mode candidates are obtained by detecting peaks on the smoothed vibration spectrum $\sigma(\omega)$. 
(b)~Modes that exhibit high spatial correlation with previously accepted ones are pruned (discarded modes and their spectral peaks are marked in red).
(c) Remaining modes are evaluated by their spatial total variation (TV).
Modes that violate the expected monotonic increase of spatial complexity with frequency are identified as outliers (cyan) and rejected.
}
\afterfigure
\label{fig:findmodes}
\end{figure}
}

\newcommand{\figGolden}{
\begin{figure}[!t]
    \centering
    \includegraphics[width=\linewidth]{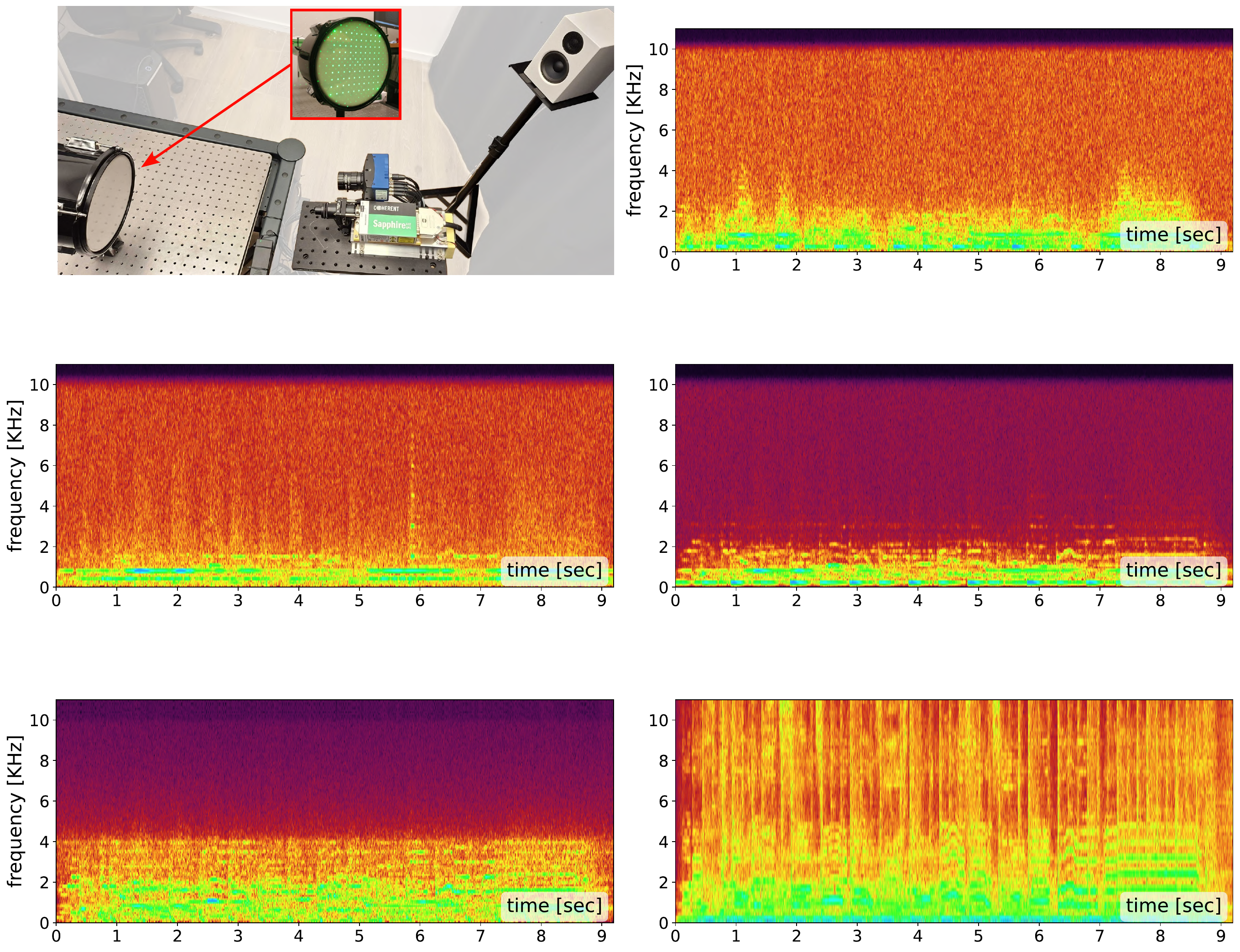}
    \put(-210, 121){\capfont experimental setup}
    \put(-95,  121){\capfont \textbf{(a)} single point (x-axis)}
    \put(-205, 57.5){\capfont \textbf{(b)} naive average}  
    \put(-92,  57.5){\capfont \textbf{(c)} delay and sum}
    \put(-215, -6){\capfont \textbf{(d)} our recovered signal}
    \put(-88,  -6){\capfont \textbf{(e)} source signal}   
    \vspace*{-2mm}
    \caption{Sound recovery from a drumhead. We capture the speckle vibrations of a drumhead excited by a speaker 
    using a $10\!\times\!10$ speckle point grid. \textbf{(a)}~The drumhead membrane has a poor and highly resonant reaction to the speaker's sound, yielding individual noisy measurements.
    \textbf{(b)}~Averaging all measured vibrations suppresses important frequencies due to their temporal misalignment.
    \textbf{(c)}~Estimating a global temporal delay between the signals prior to averaging fails to keep high-frequency modes, as different modes induce different phase delays between the points.
    \textbf{(d)}~Our result takes into account the physics of the vibrating surface and recovers higher frequencies while significantly suppressing noise.
    \textbf{(e)}~The source signal $s(t)$ played by the speaker.
    }
    \afterfigure
    \label{fig:golden}
\end{figure}
}

\newcommand{\figAllResults}{
\begin{figure}
\centering
    \begin{overpic}[width=\linewidth]{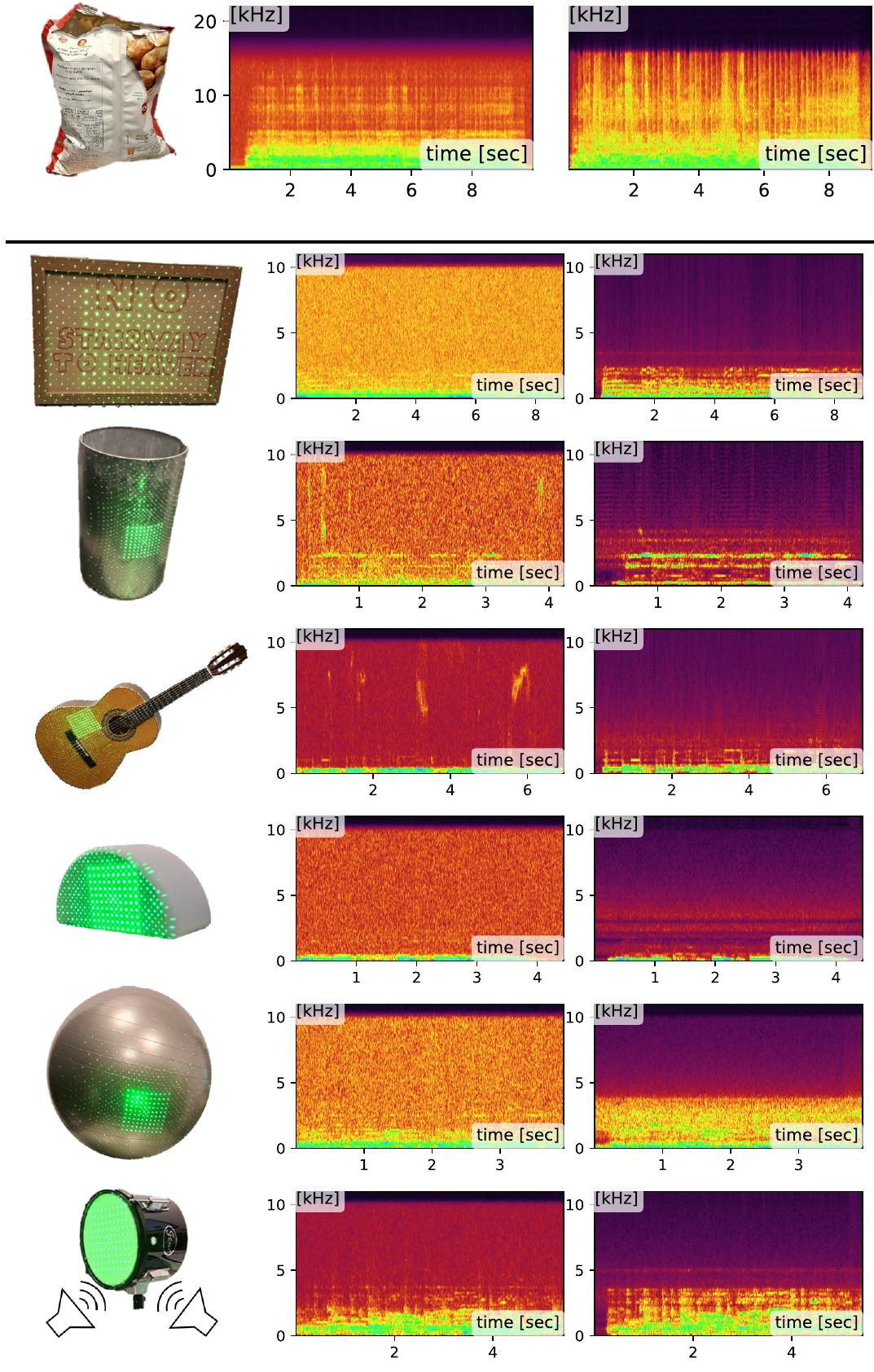}
    \put(20.5, 83.5){\capfont single point}
    \put(46.5, 83.5){\capfont source signal}
    \linethickness{1pt}
    \put(7, -1){\capfont object}
    \put(25.5, -1){\capfont single point}
    \put(51, -1){\capfont \textbf{ours}}
    \end{overpic}
    \caption{
Results across objects having various geometries and materials.
The top row (chips bag) serves as a reference-- where a single point already performs well. All subsequent objects present greater challenges due to diverse materials, thicknesses, and shapes. Despite limited or irregular surface coverage, our method yields denoised reconstructions. The final example (drum) further demonstrates robustness under dual sound sources. 
    }
    \label{fig:all_results}
    \afterfigure
\end{figure}
}

\newcommand{\figDifferentNumModes}{
\begin{figure}
    \begin{overpic}[width=\linewidth]{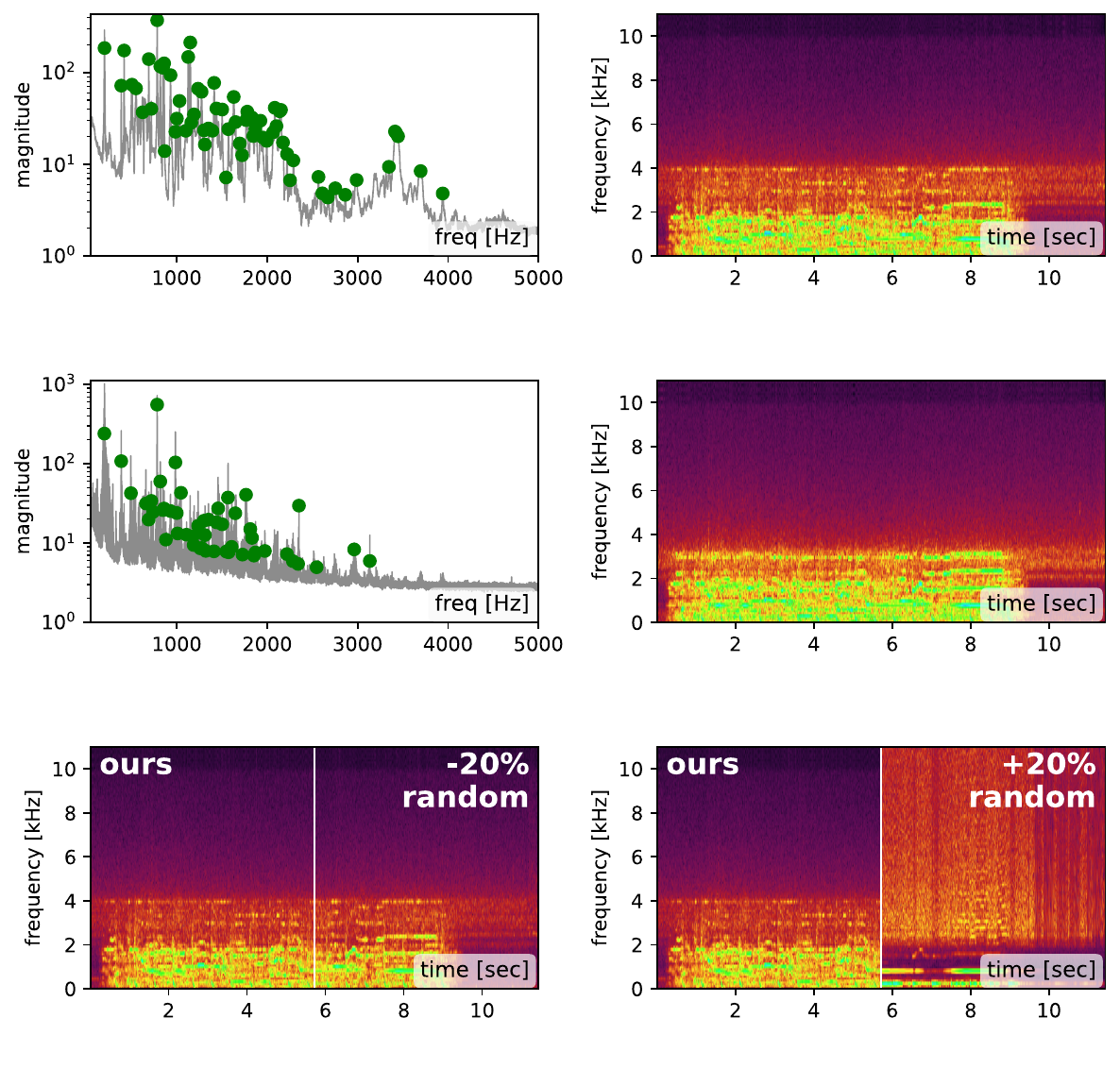}
    \put(25, 92){$\sigma\left(\omega\right)$}
    \put(25, 59){$\sigma\left(\omega\right)$}
    \put(18, 67.5){\capfont \textbf{(a)}~reconstruction using modes from a clap}
    \put(13, 34.8){\capfont \textbf{(b)}~reconstruction using modes from signal itself}
    \put(5, 1.7){\capfont \textbf{(c)}~removing (left) or adding (right) 20\% random frequencies}
    \end{overpic}
    \vspace*{-5mm}
\caption{
The experiment compares reconstructions whose mode-frequencies were obtained from different sources or from sources perturbed by random omission/addition. Inaccurate frequency detection introduces audible artifacts.
}
\label{fig:different_modes}
\end{figure}
}

\newcommand{\figCalibrated}{
\begin{figure}[t!] 
\centering
\begin{overpic}[width=\columnwidth]{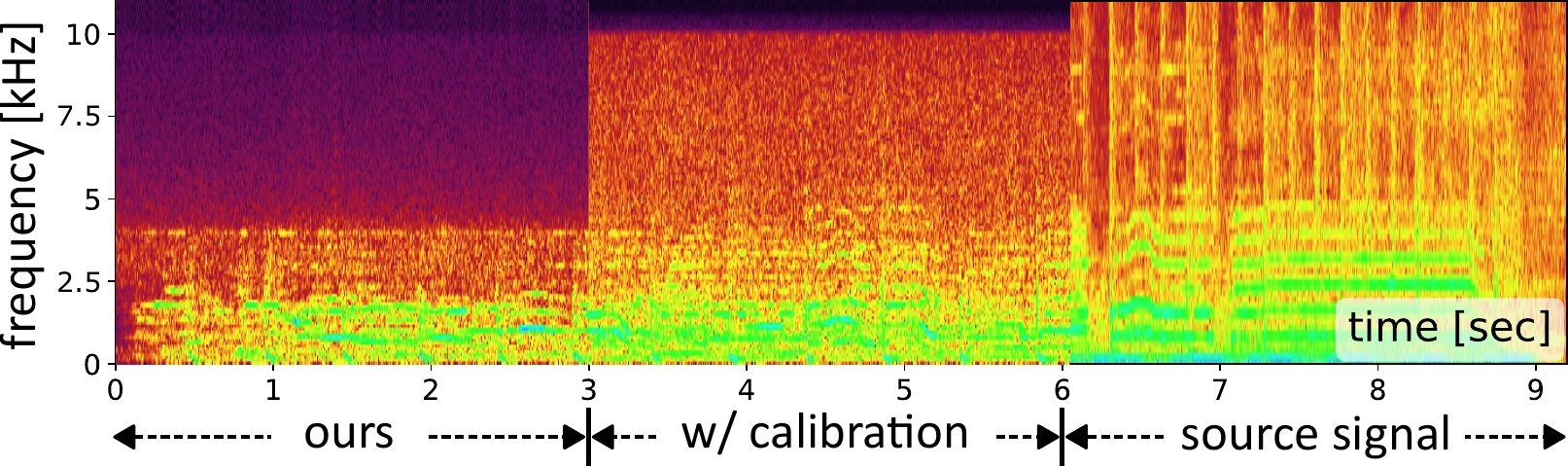} 
 \end{overpic}
\caption{Comparison between our model-based sound recovery and a calibrated baseline using a reference chirp signal. Our method reconstructs most of the frequency content with high fidelity, closely matching the calibrated result 
without using any additional intervening signals or supervision.
}
\label{fig:calibrated}
\afterfigure
\end{figure}
}

\newcommand{\figFullResults}{
\begin{figure}[h!]
\centering
    \begin{overpic}[width=\linewidth]{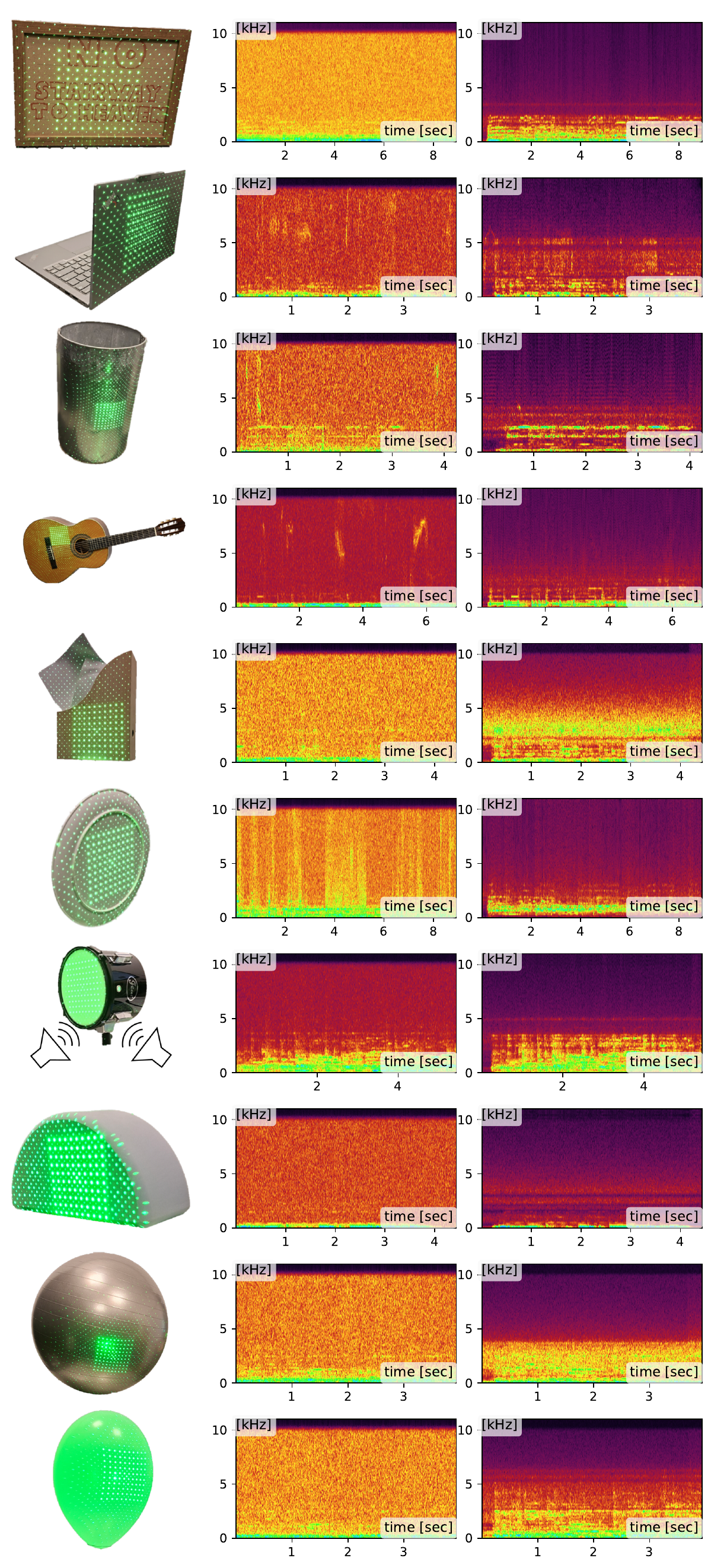}
    \put(5, -1){\capfont object}
    \put(18, -1){\capfont single point}
    \put(38, -1){\capfont \textbf{ours}}
    \end{overpic}
    \vspace*{2mm}
    \caption{
Results across objects having various geometries and materials.
All objects present challenges due to diverse materials, thicknesses, and shapes. Despite limited or irregular surface coverage, our method yields denoised reconstructions. 
}
    \label{fig:all_results_supp}
    \vspace*{15mm}
\end{figure}
}

%% file: ALL_TABLES.tex
\newcommand{\tabViSQOL}{
\begin{table}
\centering
\begin{tabular}{lllll}
\toprule
\multicolumn{5}{c}{ViSQOLAudio-NSIM$\uparrow$} \\
\midrule
Experiment & Single & Avg & DnS & Ours \\
\midrule
drum & 0.20 & 0.19 & 0.18 & \textbf{0.32} \\
frame & 0.22 & 0.15 & 0.23 & \textbf{0.46} \\
laptop & 0.28 & 0.22 & 0.33 & \textbf{0.44} \\
trash & 0.19 & 0.24 & 0.24 & \textbf{0.43} \\
guitar & 0.16 & 0.26 & 0.24 & \textbf{0.32} \\
binder & 0.26 & 0.27 & 0.37 & \textbf{0.43} \\
plate & 0.19 & 0.17 & 0.23 & \textbf{0.34} \\
drum (stereo) & 0.26 & 0.26 & 0.27 & \textbf{0.38} \\
yoga block & 0.16 & 0.08 & 0.16 & \textbf{0.29} \\
physio ball & 0.35 & 0.35 & 0.39 & \textbf{0.39} \\
balloon & 0.27 & 0.17 & 0.39 & \textbf{0.46} \\
\midrule
\textbf{Mean:} & 0.23 & 0.21 & 0.27 & \textbf{0.39} \\
\bottomrule
\end{tabular}
\captionof{table}{\textbf{ViSQOLAudio-NSIM}~\cite{hines2015visqolaudio, hines2012speech} (higher is better). The raw perceptual similarity index used inside ViSQOLAudio before Mean Opinion Score (MOS) mapping. It measures structural similarity between cochleagram patches of the reference and degraded signals and is not tied to speech-specific MOS training. Higher values indicate closer perceptual similarity.}
\label{tab:visqol-nsim}

\vspace*{10mm}

\centering
\begin{tabular}{lllll}
\toprule
\multicolumn{5}{c}{ViSQOLAudio-MOS$\uparrow$} \\
\midrule
Experiment & Single & Avg & DnS & Ours \\
\midrule
drum & \textbf{2.05} & 1.90 & 1.96 & 1.63 \\
frame & 1.22 & 1.16 & 1.31 & \textbf{1.96} \\
laptop & 1.44 & 1.61 & 1.90 & \textbf{2.06} \\
trash & 1.07 & 1.12 & 1.05 & \textbf{2.58} \\
guitar & \textbf{1.74} & 1.59 & 1.66 & 1.58 \\
binder & 1.47 & 1.54 & 1.55 & \textbf{2.45} \\
plate & 1.24 & 1.43 & 1.45 & \textbf{1.95} \\
drum (stereo) & 1.69 & 1.65 & 1.55 & \textbf{1.84} \\
yoga block & 1.16 & 1.00 & 1.19 & \textbf{2.67} \\
physio ball & 1.37 & 1.54 & 1.32 & \textbf{1.57} \\
balloon & 1.58 & 1.43 & \textbf{1.86} & 1.81 \\
\midrule
\textbf{Mean:} & 1.46 & 1.45 & 1.53 & \textbf{2.01} \\
\bottomrule
\end{tabular}
\captionof{table}{\textbf{ViSQOLAudio-MOS}~\cite{hines2015visqolaudio} (higher isbetter). Perceptual audio-quality metric based on the ViSQOLAudio model. It compares a reference and degraded signal using a cochleagram front-end and patch-based similarity, then maps the result to a Mean Opinion Score (MOS) prediction. Originally trained on general-audio listening tests, it approximates subjective quality.}
\label{tab:visqol-mos}
\end{table}
}

\newcommand{\tabSIMRSTFT}{
\begin{table}
\centering
\begin{tabular}{lllll}
\toprule
\multicolumn{5}{c}{Scale Invariant, Multi Resolution STFT distance$\downarrow$} \\
\midrule
Experiment & Single & Avg & DnS & Ours \\
\midrule
drum & 4.68 & 4.21 & 4.60 & \textbf{3.54} \\
frame & 3.02 & 3.46 & 3.15 & \textbf{1.87} \\
laptop & 4.01 & 5.17 & 4.59 & \textbf{3.88} \\
trash & 3.65 & 3.27 & 3.46 & \textbf{3.01} \\
guitar & 3.23 & \textbf{2.47} & 2.59 & 3.49 \\
binder & 3.00 & 3.51 & 3.14 & \textbf{2.72} \\
plate & 3.22 & 3.60 & 3.82 & \textbf{3.21} \\
drum (stereo) & \textbf{4.27} & 4.56 & 4.41 & 4.32 \\
yoga block & 3.92 & 4.20 & \textbf{3.66} & 3.68 \\
physio ball & \textbf{2.92} & 3.40 & 3.16 & 4.00 \\
balloon & \textbf{3.18} & 3.39 & 3.96 & 3.73 \\
\midrule
\textbf{Mean:} & 3.55 & 3.75 & 3.69 & \textbf{3.40} \\
\bottomrule
\end{tabular}
\captionof{table}{\textbf{Scale Invariant, Multi Resolution STFT distance}~\cite{yamamoto2020simrstft, steinmetz2020auraloss} (lower is better). A distance metric between recovered signal and reference source signal. The measure is a scale invariant, multi resolution STFT distance.}
\label{tab:si-mr-stft}

\vspace*{10mm}

\centering
\begin{tabular}{lllll}
\toprule
\multicolumn{5}{c}{Scale Invariant, Multi Resolution STFT distance}\\
\multicolumn{5}{c}{(Perceptual weighting)$\downarrow$} \\
\midrule
Experiment & Single & Avg & DnS & Ours \\
\midrule
drum & 3.65 & 3.39 & \textbf{3.28} & 3.39 \\
frame & 3.64 & 5.01 & 3.94 & \textbf{2.01} \\
laptop & 3.11 & 3.66 & 3.92 & \textbf{2.91} \\
trash & 3.56 & 3.57 & 3.88 & \textbf{2.61} \\
guitar & 3.45 & 2.91 & 3.04 & \textbf{2.53} \\
binder & 3.00 & 3.18 & 3.20 & \textbf{2.48} \\
plate & 3.44 & 3.57 & 4.37 & \textbf{2.70} \\
drum (stereo) & 3.67 & \textbf{3.59} & 4.16 & 3.97 \\
yoga block & 4.14 & 4.01 & 3.87 & \textbf{3.68} \\
physio ball & \textbf{3.08} & 3.22 & 3.12 & 3.35 \\
balloon & \textbf{3.00} & 3.12 & 3.42 & 3.58 \\
\midrule
\textbf{Mean:} & 3.43 & 3.57 & 3.65 & \textbf{3.02} \\
\bottomrule
\end{tabular}
\captionof{table}{\textbf{Scale Invariant, Multi Resolution STFT distance (Perceptual weighting)}~\cite{yamamoto2020simrstft, steinmetz2020auraloss} (lower is better). A distance metric between recovered signal and reference source signal. The measure is a scale invariant, multi resolution STFT distance with perceptual weighting.}
\label{tab:si-mr-stft-p}
\end{table}
}


%% file: sec/0_abstract.tex
\begin{abstract}
Optical vibration sensing enables recovering the scene sound directly from the surface vibration of nearby objects, turning everyday objects into ``visual microphones''. However, most prior methods had focused on capturing the vibrations of specific objects with highly 
favorable vibration responses. These include objects where the surface vibrations are generated by the object itself (\eg, speaker membrane or guitar body) or objects consisting of a thin membrane which is highly reactive to sound (\eg, a chip bag or the leaf of a plant).
In this paper, we tackle sound recovery for a more challenging class of solid objects whose vibration responses are poor or highly resonant. We simultaneously capture vibrations for multiple surface points on the object using a speckle-based vibrometry imaging system. Then, we derive a novel physics-guided vibration formation model that relates the scene sound source to the captured multi-point multi-axis vibrations via the object's vibrational modes. The model is then used to reverse the resonant transfer function of the vibrating object, fusing multiple vibration signals to estimate the original sound source in the scene. We evaluate our approach by recovering sound from a variety of everyday objects, demonstrating that it significantly outperforms traditional single-point speckle vibrometry in challenging scenarios and other signal-processing-based methods for multi-signal fusing.
\end{abstract}

%% file: sec/1_intro.tex
\section{Introduction}

From massive bridges to the gentle breathing of a sleeping baby, the ability to enhance tiny, imperceptible motions in the world around us has fascinated computer vision researchers for over two decades \cite{liu2005motion,wu2012eulerian,wadhwa2013phase,wadhwa2014riesz,zhang2017video,elgharib2015video,oh2018learning,bouman2013estimating,chen2024event,buyukozturk2016smaller,chen2017video,wadhwa2016eulerian,davis2015image,zhou2016vibration,chen201558}. One particularly intriguing and elusive type of motion is that of scene objects set in motion by sound within the observed scene \cite{kac1966can_you_hear_the_shape_of_a_drum,davis2014visual,Sheinin:2022:Vibration,zalevsky2009simultaneous,cai2025event2audio, nassi2020lamphone}. Recovering these vibrations, optically, holds the promise of capturing the complete audio-visual experience of a scene with nothing more than a single device -- a \textit{camera}.

Davis \etal~were the first to demonstrate a \textit{passive} approach combining a high-speed camera with motion magnification algorithms to extract scene sound from a video \cite{davis2014visual}. 
However, such passive approaches are generally impractical because the vibration amplitudes induced by sound are extremely small in standard video. In contrast, active techniques enabled the remote extraction of scene vibrations by illuminating the vibrating surface with a laser. The resulting interference pattern, called speckle, optically amplifies the surface micro-vibrations by orders of magnitude \cite{zalevsky2009simultaneous,bianchi2019long,wu2020fast,wu202120k,Sheinin:2022:Vibration,Kichler:2025,Jo_2015_ICCV,alterman2021imaging}.
However, in the context of sound recovery, these prior methods focused on capturing vibrations of objects with especially favorable vibration responses. These include `active' objects, where the surface vibrations are generated by the object itself (\eg, a speaker membrane or guitar body), or thin, membrane-like surfaces that are highly reactive to sound (\eg, a chip bag). In this work, we seek to extend vision-based sound recovery to a class of more challenging objects whose impulse responses to ambient sounds are either weak or highly resonant. 


Our approach relies on recent works that enable a single camera to simultaneously record vibrations at multiple scene points \cite{zalevsky2009simultaneous,Sheinin:2022:Vibration,Kichler:2025,Zhang:2023,cai2025event2audio,Smith2017,Smith_2018_CVPR}.
These advances naturally raise the question: Can we combine the noisy vibrations from \textit{many surface points captured on a single object} to yield a single, maximally denoised sound measurement?
The answer is non-trivial, as underscored by the poor performance of naive signal averaging (see Fig.~\ref{fig:teaser}).
Moreover, unlike in a microphone array, where a global time delay relates all pairs of individual measurements \cite{Benesty2017}, our measurements are linked by the complex mechanical surface vibrations induced by the impinging acoustic pressure wave. Furthermore, each speckle-vibration measurement yields two-axis surface tilts that may have different acoustic profiles \cite{Sheinin:2022:Vibration}. Even for the single-point case, none of the works mentioned above provided a principled solution for merging two-axis vibration, resorting instead to either playing one axis \cite{Sheinin:2022:Vibration,zalevsky2009simultaneous} or assuming a global delay, as in two independent microphones \cite{cai2025event2audio}. Starting from first principles, we derive a new forward model linking the mechanical surface vibrations induced by scene sound to the resulting 2D speckle shifts, enabling the fusion of 
signals from \textit{multiple} 2D surface points in a principled, physics-guided manner.

The key challenge in fusing multi-point signals is estimating and correcting their mutual \textit{frequency-dependent} phase delays and magnitude differences before averaging, without overfitting the noise frequencies. To tackle this, we use the insight that the object's modal frequencies and shapes, if known, provide this missing link. Moreover, for most objects, the modes form an orthogonal basis spanning all surface vibrations. Therefore, the modes can be used to approximately invert the object's spatio-temporal impulse response and estimate the underlying sound that induced the vibrations. 
Fig.~\ref{fig:teaser} provides an overview of our framework.
Firstly, we identify the mode shapes and frequencies from the measured vibrations. Then, we model the captured speckle shifts using this overcomplete modal basis and estimate the underlying scene sound by inverting the model. Thus, to the best of our knowledge, we introduce the first method for vision-based sound recovery that ties speckle-based vibrometry with the mechanical vibrations of everyday objects. 

We evaluate our method by capturing sound from various everyday objects having different materials and shapes (\eg, a laptop, binder, guitar, drum, picture frame, and more). Our method yields significantly superior sound recoveries with respect to a single point capture \cite{Sheinin:2022:Vibration, zalevsky2009simultaneous,wu2020fast,bianchi2019long,cai2025event2audio}. Moreover, we show that our method outperforms `classical' signal-processing-based approaches such as averaging and delay-and-sum in denoising the captured sound and `equalizing' its acoustics to remove the resonant timbre induced by the object's modes. We believe that our method constitutes a significant advancement in visual sound recovery, moving the field closer towards a general-purpose, camera-based acoustic sensing of real-world scenes.

%% file: sec/2_background.tex

\section{Background}
\subsection{Vision-based scene sound recovery}
\label{sec:speckle}
In the computer vision community, extracting sound from videos first evolved as a special case of scene motion magnification \cite{liu2005motion,wu2012eulerian,wadhwa2013phase,wadhwa2014riesz,oh2018learning,wadhwa2016eulerian}. 
These methods used high-speed cameras to capture vibrating surfaces and estimate sound from texture displacements. 
Earlier, in the optics domain, sound recovery from videos relied on an active approach called
 speckle vibrometry, which senses reflected laser interference patterns (\ie, \textit{speckle}) \cite{zalevsky2009simultaneous,wu2020fast,wu202120k}. In this method, surface vibrations are derived from shifts in defocused speckle patterns, originating from surface microstructures, making it far more sensitive to small vibrations than texture-based methods \cite{zalevsky2009simultaneous}. Sheinin \etal~introduced a dual-camera system eliminating the need for high-speed capture while enabling multi-point scene acquisition \cite{Sheinin:2022:Vibration}. Mingxuan \etal~then extended the visual vibrometry to event cameras \cite{cai2025event2audio,Howard2023,Niwa_2023_CVPR,Zhou_2025_ICCV}.
Recently, Kichler \etal~extended speckle vibrometry to a 2D scene grid to recover container vibrations and infer their liquid contents. In this paper, we leverage the 2D scene speckle vibration recovery and focus on the novel task of modeling and estimating the underlying sound that yielded the vibrations.

\subsection{Modeling surface vibrations}
\label{sec:vibrations_background}
The dynamics of a thin, linearly elastic surface~-- whether a \emph{membrane} dominated by in-plane tension or a \emph{plate} dominated by bending stiffness~-- can be described by the general wave equation \cite{rao2007continuous, timoshenko1959}:
\begin{equation}
\rho \frac{\partial^2 u}{\partial t^2} + c(\mathbf{x})\frac{\partial u}{\partial t} -   T \nabla^2 u + D \nabla^4 u = f(\mathbf{x},t),
\label{eq:membrane_vib}
\end{equation}
where $u(\mathbf{x},t)$ is the out-of-plane displacement at surface location $\mathbf{x}$, $\rho$ is the areal mass density, $c(\mathbf{x})$ is the material damping coefficient, $T$ is the in-plane tension (for membranes), $D$ is the flexural rigidity (for plates), and $f(\mathbf{x},t)$ is the external forcing per unit area.  
For membranes $D\!=\!0$, and for thin plates under negligible tension $T\!=\!0$. 
Boundary conditions determine the resulting modal structure.

Under linear vibration assumptions, the surface displacement can be expanded as a sum over mode shapes \cite{rao2007continuous}:
\begin{equation}
    u(\mathbf{x},t) = \sum_{k=1}^{K}\phi_k(\mathbf{x})\, q_k(t),
    \label{eq:displace}
\end{equation}
where $\phi_k(\mathbf{x})$ are the spatial eigenfunctions (mode shapes) and $q_k(t)$ are the temporal modal coordinates. 

Projecting Eq.~\eqref{eq:membrane_vib} onto each mode and invoking orthogonality yields a second-order differential equation of motion for each modal coordinate \cite{cremer2005, rao2007continuous}.
Assuming also that the driving pressure field is approximately uniform across the surface, \ie, $f(\mathbf{x},t)\!=\!p(t)$ yields:
\begin{equation}
    \ddot{q}_k(t) + 2 \zeta_k \omega_k \dot{q}_k(t) + \omega_k^2 q_k(t)
    = \alpha_k p(t) + \eta_k(t),
    \label{eq:diff_2nd_order}
\end{equation}
where \( \omega_k \) is the natural frequency of mode \( k \), \( \zeta_k \) is its damping ratio, \( \alpha_k \) is the modal coupling coefficient to the uniform pressure field \(p(t)\), and \( \eta_k(t) \) represents measurement or modeling noise.  
The values of \( \omega_k \), \( \zeta_k \), and \( \alpha_k \) depend on the material properties (\eg, \( \rho \), \( T \), \( D \)), geometry, and boundary conditions of the surface.

In the frequency domain, Eq.~\eqref{eq:diff_2nd_order} defines an analytic relationship between the driving sound signal and the mechanical response of each mode:
\begin{align}
    Q_k(\omega) &=  G_k(\omega)\, P(\omega),\label{eq:beforeG}  \\
    G_k(\omega) &= \frac{\alpha_k}{-\omega^2 + j\,2 \zeta_k \omega_k \omega + \omega_k^2}.\label{eq:G}
\end{align}
The function $G_k(\omega)$ is the transfer function of a second-order linear oscillator \cite{ewins2000}, representing the frequency response of the $k$-th mode.

Finally, under the same assumptions, Eq.~\eqref{eq:displace} can be rewritten 
as \cite{rao2007continuous}:
\begin{equation}
    u(\mathbf{x},t) =  \sum_{k=1}^{K} \phi_k(\mathbf{x})\, \big(p(t) * g_k(t)\big),
    \label{eq:general_conv}
\end{equation}
where $g_k(t)$ is the impulse response associated with mode $k$, whose Fourier transform is $G_k(\omega)$.  
This expression holds generally for any linear vibrating surface. In simple cases such as a uniform membrane or thin plate, $g_k(t)$ can be derived analytically from Eq.~\eqref{eq:G}; for more complex or spatially varying structures, $g_k(t)$ must be obtained numerically or empirically.

%% file: sec/3_recover_sound.tex
\newcommand{\parvspace}{\vspace*{-6pt}}
\section{Recovering sound from surface vibrations}
\label{sec:inverse}
\subsection{Problem overview}
Our camera captures the vibrations of a scene object on a 2D grid of surface points.
As detailed below, each measured point yields a two-dimensional signal describing the local surface vibrations. Our goal is to fuse all point signals into a single, denoised audio estimate.
Broadly speaking, yielding the denoised audio requires knowing the transfer function between the sound source and each 1D measurement stemming from a surface point and vibration axis.    

%
%
However, unlike a microphone array, our point signals are not linked by simple time-of-arrival delays but by the surface’s mechanical wave propagation.
Since sound travels much faster in solids \cite{Borg2025}, the mechanical vibrations dominate the object's overall motion.
Fig.~\ref{fig:signals}(b) shows this effect,
where two surface points (red and green) 
exhibit different phase delays for different frequency components of a single recording.
Consequently, simple averaging, or even beamforming techniques such as delay-and-sum, yield degraded results, as some frequencies interfere constructively while others are canceled out~\cite{Wu:25}.

\figSignals

A second challenge arises because different measurement points exhibit distinct spectral energy distributions, depending on their geometric locations relative to the mode shapes. For example, points near a modal peak or valley contribute little energy around that mode’s frequency, since our vibration measurement amplitudes are proportional to the mode’s spatial gradient, which vanishes at extrema (see Fig.~\ref{fig:signals}\textbf{(c)}). Moreover, the spectrum across all points may be dominated by modal frequencies, imparting an unnatural resonant timbre that we aim to suppress.

In Sec.~\ref{sec:vibration_to_speckle}, we derive a forward model that links the measured speckle shifts to the underlying scene sound via the surface’s modal vibrations. This model provides the missing bridge between frequency-dependent signals across the grid. Then, Sec.~\ref{sec:without_calib} shows how to estimate the underlying signal by implicitly inverting it, using an estimate of the object's modal frequencies. 
\subsection{From scene sound to speckle-shift vibrations}
\label{sec:vibration_to_speckle}
Let $s(t){\in}[-1,1]$ denote the dimensionless sound signal emitted by a source located near the imaged surface. The sound pressure driving the surface vibrations, expressed in units of \SI{}{\pascal}, is $p(t) = \gamma\, s(t)$,
where the constant $\gamma$ accounts for factors such as the source amplitude and the acoustic attenuation between the source and the surface.

Our camera recovers the two-axis image-domain speckle shifts $\mathbf{v}(\mathbf{x_n},t)\in \mathbb{R}^2$ for $N$ surface grid points, where $n\!=\!1,\ldots,N$. We use subscripts to denote the different vector dimensions; namely, $\mathbf{v}_1$ and $\mathbf{v}_2$ denote the x- and y-axis vibrations, respectively.
We assume all $N$ points fall on the surface of a single object.
The measured shifts $\mathbf{v}(\mathbf{x_n},t)$ correspond to the surface gradients up to a scaler $\beta$
\begin{equation} 
    \mathbf{v}(\mathbf{x}_n,t) = \beta \nabla_{\mathbf{x}} u(\mathbf{x}_n, t),
    \label{eq:shifts}
\end{equation}
where $\nabla_{\mathbf{x}}$ denotes the spatial gradient of $u(\mathbf{x}, t)$ \cite{Zhang:2023}.

Substituting Eq.~\eqref{eq:general_conv} and $p(t)$ into Eq.~\eqref{eq:shifts} yields the relation between our raw vibration measurements and the original sound source:
\vspace*{-2mm}
\begin{equation}
    \mathbf{v}(\mathbf{x}_n,t) \approx \gamma\beta
                        \sum_{k=1}^{K} \nabla\phi_k(\mathbf{x}_n)\, \big(s(t) * g_k(t)\big) + \eta(\mathbf{x}_n,t),
    \label{eq:main_relation}
\end{equation}
where $\eta(\mathbf{x}_n,t){\in} \mathbb{R}^2$ is some signal-independent noise incurred at point $n$.
This is our forward model, approximating the relation between scene audio $s(t)$ to the measured surface vibrations $\mathbf{v}(\mathbf{x}_n,t)$.


\subsection{Recovering scene sound}
\label{sec:without_calib}
Recovering $s(t)$ using Eq.~\eqref{eq:main_relation} requires estimating two main surface properties: (a)~the mode shape gradients at our measurement points $\modeShapeGrad$ and (b)~their corresponding impulse responses $\modeImpulseRes$. 
We approximate the latter using its analytical form given by Eq.~\eqref{eq:G}, which has three unknowns: $\alpha_k,\zeta_k,\omega_k$. Next, we describe how to estimate these parameters from the data and use them to recover $s(t)$.
 
\vspace{-5pt}
\paragraph{Estimating mode frequencies and shape gradients.} 
Here, we draw on classical modal analysis methods that use spectral analysis to detect and extract mode shapes and frequencies \cite{HE2001159}. Specifically, let
\begin{equation}
    \mathbf{V}(\mathbf{x}_n,\omega) \equiv \mathcal{F}\{\mathbf{v}(\mathbf{x}_n,t)\},
    \qquad \mathbf{V}(\mathbf{x}_n,\omega) \in \mathbb{C}^2.
\end{equation}
be the Fourier transform of point $\mathbf{x}_n$. Then the mode frequencies $\omega_k$ can be identified by finding the local peaks in the spectrum magnitude $|\mathbf{V}(\mathbf{x}_n,\omega)|$. Since we have a plurality of points, we estimate the mode shapes using
\begin{equation}
    \{\hat\omega_k\}_{k\!=\!1\ldots{}K} \leftarrow {\rm robustfindpeaks}\left(
    |\mathbf{V}_n(\mathbf{x}_n,\omega)|
    \right),
    \label{eq:find_peaks}
\end{equation}
where ${\rm robustfindpeaks}()$ is a robust procedure to identify the mode frequencies and is described in Sec.~\ref {sec:find_modes}. 

Once the mode frequencies are identified, the mode shape derivatives are extracted directly using~\cite{brincker2000modal,brinker_carlos_2015_modal_ana}:
\begin{equation}
    \nabla\hat\phi_k(\mathbf{x}_n) =
    \operatorname{Re}\left\{
    \frac{\mathbf{V}(\mathbf{x}_n,\hat\omega_k) \cdot
    \mathbf{V}_1(\mathbf{x}_0,\hat\omega_k)^{*}}
    {\mathbb{E}_{n,a}[|\mathbf{V}(\mathbf{x}_n,\hat\omega_k)|]\cdot
    \left|\mathbf{V}_1(\mathbf{x}_0,\hat\omega_k)\right|}
    \right\},
    \label{eq:mode_shapes_recover}
\end{equation}
where $\mathbb{E}_{n,a}$ is the mean operator over all points and axes, and $\operatorname{Re}\left\{\right\}$ is the real-part operator. Intuitively, the mode shape gradients are determined by the relative two-axis magnitudes (sign included) of the 
harmonic signals for each point $\mathbf{x}_n$ at the mode frequency $\omega_k$, which directly follow from $\mathbf{V}(\mathbf{x}_n,\hat\omega_k)$. The mode shapes are normalized by shifting all phase components 
to have a zero phase (using point $\mathbf{x}_0$) 
and then dividing by the mode's mean magnitude. 

\vspace{-5pt}
\paragraph{Recovering the latent sound signal.}
\label{sec:recoveringSt}
Given the estimated mode shape gradients and frequencies and Eq.~\eqref{eq:main_relation}, we recover the sound
$\hat s^{\rm inv}(t)$ by solving:
\vspace*{-1mm}
{\small
\begin{align}
    \underset{s(t),\alpha_k}{\argmin{}}\!\!
    \left\|
    \mathbf{v}(\mathbf{x}_n,t) - \sum_{k=1}^{K} \nabla\hat\phi_k(\mathbf{x}_n)\, \big(s(t) * \hat g_k(t)\big)
    \right\|_2^2
    \!+\! 
    \lambda\left\| \dot{s}(t) \right\|_2^2,
    \label{eq:invS}
\end{align}
}where $\hat g_k(t)$ is computed by $\hat{g}_k(t) {\leftarrow} \texttt{iFFT}(G_k(\omega))$ using its analytic form (Eq.~\eqref{eq:G}). We set $\zeta_k$ to some constant value in $G_k(\omega)$ while $\alpha_k$ is jointly optimized in \cref{eq:invS} (implicit in $G_k(\omega)$). 
Optimizing Eq.~\eqref{eq:invS} reverses the surface's resonant effect while also denoising the signal by taking into account the frequency-dependent phase relationship between the $N$ measurement points through the mode shapes. The result is an `equalized' signal with a `flatter' spectrum that is closer to $s(t)$ than any of the original signals.

\vspace{-10pt}
\begin{remark}
In \cref{eq:shifts}, we assume that all speckle-based shift measurements share the same optical transfer factor $\beta$, which is approximately the case in our imaging scenarios.
In the general case, however, effects like a high variance between the points' axial camera distance,
may yield different per-point scalers $\beta_n$. Nevertheless, the procedure described in Sec.~\ref{sec:recoveringSt} still holds true since the factors $\beta_n$ will simply skew the numerically recovered mode shape gradients in \cref{eq:mode_shapes_recover} by a constant per-point factor, while \cref{eq:invS} is agnostic to the true physical mode shape gradients of the surface. See the supplementary for a detailed derivation.
\end{remark}

%% file: sec/4_robust_mode_estimation.tex
\vspace{-5pt}
\section{Robust mode estimation}
\label{sec:find_modes}
Our method relies on recovering the modal frequencies and shapes of the observed object.\footnote{In practice, we only recover a segment of the modes' shape, defined by the positions of our laser grid points.}
As in prior modal analysis works, the modes can be extracted directly from the data, assuming their frequencies are prominent in the signal’s spectrum~\cite{brincker2000modal,brinker_carlos_2015_modal_ana}. 
In practice, however, most modes are only activated by broadband excitations. We therefore assume that, \textit{in aggregate}, over the span of a longer recording (1min, 5min, or even 30min), some incidental broadband events, such as snaps, claps, or impacts, occur and effectively excite the full set of modes. This assumption is analogous to standard audio recording practice, in which a brief segment of ambient sound is intentionally captured to later aid noise removal. 
Thus, 
our task boils down to detecting the correct mode frequencies $\{\omega_k\}$ from the input signals. 

\figFindModes

To robustly identify the mode frequencies, we compute the standard deviation of the FFT magnitudes across all $\mathbf{x}_n$: 
\vspace*{-2mm}
\begin{equation}
\sigma(\omega) = \operatorname{std}_n \left( \, |\mathbf{V}_n(\mathbf{x}_n, \omega)| \, \right),
\label{eq:std_spectrum}
\end{equation}
The 
std $\sigma(\omega)$ helps reveal the modes because, on average, frequencies belonging to modes will have variation between the different surface points (mode-shape peaks or valleys) while uncorrelated noise has slight variation.  
We then apply a three-stage procedure that progressively enforces spectral, spatial, and physical consistency.
First, we smooth $\sigma(\omega)$ with a local \SI{5}{Hz} Savitzky–Golay filter~\cite{savitzky1964smoothing} and use SciPy’s \texttt{find\_peaks}~\cite{find_peaks_2020scipy} to extract initial mode frequency candidates (Fig.~\ref{fig:findmodes}(a)).
At this stage, all prominent spectral peaks are retained, even if some correspond to noise or redundant harmonics.
Second, we recover the corresponding mode shapes using Eq.~(\ref{eq:mode_shapes_recover}) and prune highly correlated ones (Fig.~\ref{fig:findmodes}(b)).
Finally, we compute the total variation (TV) of each mode shape,
\begin{equation*}
    \text{TV}\left(\nabla\phi_k\right) = \sum_{x,y}
    \left\|\frac{\partial}{\partial x}\nabla\phi_k\left(x,y\right)\right\|_2 +   
    \left\|\frac{\partial}{\partial y}\nabla\phi_k\left(x,y\right)\right\|_2
    \label{eq:mode_total_variation}    
\end{equation*}
and discard outliers that violate the expected monotonic increase of spatial complexity with frequency, a fundamental property of vibration eigenmodes~\cite{meirovitch2001fundamentals, ewins2000} (Fig.~\ref{fig:findmodes}\textbf{(c)}).
This final stage rejects modes exhibiting unusually high spatial frequency content at low temporal frequencies, which are typically numerical artifacts.

%% file: sec/4p5_calibrated.tex
\section{Calibration-based optimal recovery baseline}
\label{sec:calibrated}
The method described in Sections \ref{sec:inverse} and \ref{sec:find_modes} reconstructs the response of the observed surface based on the analytical model provided by Eq.~\eqref{eq:main_relation} and requires no direct access to or intervention within the scene. The frequency response of most everyday materials tends to decrease in magnitude with increasing frequency, as is evident in all our recovered results (when compared to the source signal). A natural question arises: how far are our results from the “optimal” recovery -- namely, the best recovery possible given the object's transfer function?

To answer this question, we propose a procedure to estimate the transfer function between the sound source and each 1D measurement from $\mathbf{v}(\mathbf{x}_n,t)$ directly, thereby bypassing our analytical model entirely. This can be achieved by playing a known reference \textit{calibration} sound in the scene, $s^{\rm ref}(t)$, while recording its resulting vibrations $\mathbf{v}^{\rm ref}(\mathbf{x}_n,t)$. The known reference signal allows us to invert the response at each point independently, then average the inverted signals. This procedure, described next, is inspired by modern microphone beamforming techniques \cite{Benesty2008,Benesty2017,Haykin1984,Sullivan2008,Richards2014,Cohen2010}, and provides a good proxy for the best-case benchmark of signal recovery 
(see Fig.~\ref{fig:calibrated}).


Given a known reference signal $s^{\rm ref}(t)$, Eq.~\eqref{eq:main_relation} simplifies to a convolutional relation between each individual vibration component and the reference signal:
\begin{equation}
\mathbf{v}_a^{\rm ref}(\mathbf{x}_n,t) \approx s^{\rm ref}(t) * h_{n,a}(t) 
\label{eq:conv_relation}
\end{equation}
where $h_{n,a}(t)\in \mathbb{R}^{1 \times T{\rm conv}}$ denotes an impulse response of point $n$ at vibration axis $a$ with $T_{\rm conv}$ taps. Using the reference signal, we then estimate and store the individual inverse filters $\hat{h}_{n,a}(t)^{-1}$ by solving the following least-squares problem:
{\small
\begin{equation}
\hat{h}_{n,a}(t)^{-1} = \underset{h(t)}{\rm argmin}\left\|\mathbf{v}_a^{\rm ref}(\mathbf{x}_n,t)* h(t) - s^{\rm ref}(t)\right\|_2^2.
\label{eq:lsq_for_filt}
\end{equation}
}
Then, for every new measurement, we estimate a denoised, demodulated signal by applying the calibrated inverse filter to each vibration channel and averaging the results:
\vspace*{-1mm}
\begin{equation} 
\hat{s}^{\rm new}(t) = \frac{1}{2N}\sum_{\rm a=1}^2\sum_{n=1}^N \mathbf{v}^{\rm new}_a(\mathbf{x}_n,t)*\hat{h}_{n,a}(t)^{-1} 
\label{eq:avg_align} 
\end{equation}

%% file: sec/5_implement_details.tex
\section{Implementation details}

\paragraph{Capturing speckle vibrations:} We recreated the experimental setup of Kichler~\etal~\cite{Kichler:2025} while creating a $10\!\times\!10$ dot grid using a HOLOEYE beam splitter \cite{HOLOEYE_DE-R351}. The camera was set to \SI{22000}{fps}, while only reading out 10 regions of interest, each 10 pixels high. Only for \cref{fig:all_results}(Top row), the camera speed was \SI{44100}{fps}.
The two-axis speckle shifts were computed using the PCLK+ algorithm \cite{Kichler:2025}.
The resulting shift signals were preprocessed using a $7^{th}$-order Butterworth BPF (\SI{50}{}–\SI{10000}{Hz}), and all subsequent computations were performed on these filtered signals.

\paragraph{Recovering sound via optimization:} 
We set $\zeta_k\!=\!0.01$ in all experiments and used PyTorch to solve the optimization of Eq.~\eqref{eq:invS}.  Specifically, we performed gradient descent using Adam~\cite{kingma2017adam} for 10,000 steps with a fixed learning rate of $10^{-4}$ with $\lambda\!=\!1$. At each iteration, we computed the loss for all grid points $\mathbf{x}_n$ and all time steps $t$. Recovery took about \SI{27}{\second} on an NVIDIA RTX 4090 GPU.
In Section~\ref{sec:calibrated}, we find the inverse filters using 
PyTorch's least squares solver 
\texttt{lstsq}, where the filter size was set to 4096 taps. 

%% file: sec/6_experimental_evaluation.tex
\section{Experimental results}
\label{sec:results}
\figGolden
\figAllResults
We extensively tested our approach across a variety of objects and diverse audio signals. 
\textbf{Audio results are available on the website.}
In our experiments, we placed each object approximately \SI{50}{\centi\meter} from the camera, resulting in a dense grid of measurements captured on the object’s surface (see Figs.~\ref{fig:golden} and \ref{fig:all_results}).
A speaker was then positioned approximately \SI{150}{\centi\meter} away, and short music excerpts from various songs were played while the camera recorded the resulting speckle vibrations. The speaker was intentionally configured to moderate-to-low loudness, such that the resulting sound from a single vibration measurement was noise-limited, with a signal-to-noise ratio close to unity. 
For each object, we additionally recorded a single clap sound, manually produced by a person randomly positioned between \SI{60}{\centi\meter} and \SI{150}{\centi\meter} from the object.
Unless stated otherwise, the clap recordings served as the reference input for the mode extraction process described in Section~\ref{sec:find_modes}.
Finally, in several comparison figures, we show the spectrograms of the source signal before playback, called the ``source signal.” Note that these spectrograms do not include the effects of propagation through the speaker-room system and thus serve as an idealized reference for the captured sound.

\vspace{-7pt}
\paragraph{Comparison to alternative signal fusion techniques}
The top row of Fig.~\ref{fig:all_results} shows that the audio recovered from a single point (y-axis shift) on a chip bag has predictably excellent quality due to the bag's favorable material.
In contrast, Fig.~\ref{fig:golden}(a) demonstrates that under the same conditions (\ie, volume and distance to the source), a single-point measurement from a drumhead is both noisy and heavily ``colored" by the drum’s resonant frequencies, resulting in a strongly drum-like timbre in the recovered sound.
Averaging signals across all points yields even worse-sounding results, as random phase cancellations suppress important frequency components (Fig.~\ref{fig:golden}(b)). Fig.~\ref{fig:golden}(c) shows the result when a global optimal time delay between all point signals is estimated and compensated before averaging \cite{Benesty2017}. This classical beamforming approach effectively reduces high-frequency noise but still fails to reconstruct the full spectral content of the signal, since using a \emph{single} constant delay aligns only the dominant low-frequency modes, leaving many higher frequencies lost. Our recovery, shown in Fig.~\ref{fig:golden}(d), yields a substantially improved reconstruction that is denoised, spectrally richer, and perceptually less affected by the drum’s resonant modes. 
Fig.~\ref{fig:all_results} shows additional selected results. 
Extensive quantitative evaluation is provided in the supplementary. 
Our method performs well across objects with diverse materials (wood, metal, plastic, rubber), geometries (planar, curved), and irregular shapes (such as the wooden binder and guitar body). 
The approach even works for solid objects like the Yoga block in Fig.~\ref{fig:all_results}, despite not being optimized for volumetric structures.


\figDifferentNumModes
\vspace{-7pt}
\paragraph{Testing different mode extractions}
Fig.~\ref{fig:different_modes} shows how different mode extraction sources affect recovery 
for the drum experiment. 
Fig.~\ref{fig:different_modes}(a)~corresponds to modes extracted from a clap recording. In contrast, Fig.~\ref{fig:different_modes}(b) uses modes extracted directly from the original recording. Both reconstructions exhibit comparable quality and similar acoustic characteristics, demonstrating that the signal itself can serve as a reliable source for mode extraction.
We further examined the influence of mode frequency accuracy on reconstruction quality. Fig.~\ref{fig:different_modes}(c) shows that randomly omitting 20\% of our recovered mode frequencies slightly reduces sharpness but preserves the overall timbre, whereas adding 20\% spurious frequencies introduces pronounced artifacts and unnatural resonances. 
In other words, having the right frequencies matters more than having all of them.

\vspace{-7pt}
\paragraph{Comparison to calibrated baseline}
Fig.~\ref{fig:calibrated} compares our recovery with a calibrated recovery. Here, we repositioned the speaker to the opposite side of its position in Fig~\ref{fig:golden}. Then, we repeated the `Golden' recording while also recording a logarithmic chirp signal spanning from \SI{50} to \SI{10}{\kilo \Hz}.
First, we recover the signal using our method and the modes extracted in Fig.~\ref{fig:golden}, yielding the signal in Fig.~\ref{fig:calibrated} (left). We then set the chirp audio file as $s^{\rm ref}(t)$ and use Eq.~\eqref{eq:lsq_for_filt} to compute inverse filters, which are applied to the same recording to produce the result in Fig.~\ref{fig:calibrated} (middle). Fig.~\ref{fig:calibrated} shows that our method reconstructs most frequencies, though the highest-frequency modes are underrepresented.
\figCalibrated

%% file: sec/7_disscusion.tex
\section{Discussion and limitations}
\label{sec:discuss}
\noindent\textbf{Mode-based sound extraction:}
Our method relies on a simplified linear, mode-based model of surface vibrations. Real objects may depart from this idealization due to spatially non-uniform forcing, complex geometry, heterogeneous material properties, boundary effects, and variations in the optical measurement process (\ie, variations in $\beta$). Nevertheless, our experiments indicate that the model captures the dominant vibration structure sufficiently well to yield strong and consistently superior sound recovery in practice. The recovered modes $\nabla\hat\phi_k(\mathbf{x}_n)$ should therefore be interpreted as empirical effective modes measured from sampled surface points rather than exact analytical eigenfunctions of the motion PDE. In this sense, they implicitly absorb some of the physical factors listed above.


\noindent\textbf{Insights from experiments:}
Our experiments reveal several interesting factors. Firstly, since our method relies on mode extraction, it performs best on stiff objects exhibiting a `sharp knock sound' upon knocking (simply put: if it knocks -- it works). Softer objects, such as a loose canvas painting, primarily exhibit low-frequency modes, which degrade performance.
The spacing and coverage of the laser point grid relative to the object surface also influences performance. When the grid spans a larger portion of the vibrating surface (Fig.~\ref{fig:golden}), the individual point signals exhibit substantial phase variation across the dominant modes, thereby making our frequency-dependent phase alignment approach yield notably superior results. Conversely, if the points’ spatial baseline is small relative to the wavelength of the dominant standing-wave modes (guitar in Fig.~\ref{fig:all_results}), the resulting phase variation between signals is small, making the setup better suited to simple averaging.
However, larger baselines on non-planar objects may degrade the captured speckle vibrations in our current camera due to geometric distortions of the 2D speckle grid on the camera's sensor.
We also examined excitation by multiple sound sources. In the drum experiment of Fig.~\ref{fig:all_results}, two speakers were placed on opposite sides of the drum and driven with a stereo signal having distinct left and right channels. The recovered sound includes contributions from both channels, though one is more prominent, likely due to unequal speaker distances, since sound intensity in air follows an inverse-square law.

\noindent\textbf{Limitations of mode extraction and opportunities for improvement:}
Our reconstruction quality relies on the correctness of the extracted modes. Our results show that our mode-extraction process mostly identified modes with frequencies below \SI{5000}{Hz}. Higher-frequency modes are more challenging to detect due to their lower vibration amplitudes. Moreover, their spatial frequency may exceed the Nyquist frequency of our spatial laser grid, violating the TV assumption of Eq.~\eqref{eq:mode_total_variation} and making their detection even harder. Additionally, Fig.~\ref{fig:different_modes} shows that incorrectly identified modes degrade the reconstruction quality.
Thus, more sophisticated future methods for recovering vibration modes could further improve reconstruction quality within our framework. Fig.~\ref{fig:calibrated}(Middle section) supports this claim, showing that higher-frequency content is indeed present in the raw data and can, in principle, be extracted.

\noindent\textbf{Societal impact:}
Optical vibration sensing can pose 
privacy risks, as these systems can remotely capture speech from object vibrations. However, our method uses 
noticeable active illumination, and sensitive areas can be protected with absorptive or specular materials or active shakers.

%% file: sec/8_conclusion.tex
\section{Conclusions}
\label{sec:conclude}
We introduced a physics-guided framework that advances vision-based sound recovery from single-point measurements to the principled fusion of many. By linking optical speckle shifts to an object’s modal vibrations, our method transforms a set of noisy, local surface motions into a coherent, higher-fidelity reconstruction of the underlying sound. This transition--from single-point to multi-point, model-based fusion--marks a fundamental step forward, unlocking new possibilities for robust, camera-based acoustic sensing across diverse materials and scenes. In doing so, we take speckle-based audio recovery to the next level, laying the groundwork for a new generation of methods that treat everyday objects not as passive reflectors of light, but as rich, distributed microphones waiting to be heard.

%% file: sec/X_suppl.tex
\section{Spatially varying optical transfer and mode shape estimation}

In Sec.~3.2 of the main manuscript, we simplified the relationship between the measured speckle shifts $\mathbf{v}(\mathbf{x}_n,t)$ and the surface gradients $\nabla_{\mathbf{x}} u(\mathbf{x}_n, t)$ by assuming a global optical transfer factor $\beta$. However, in a general imaging scenario, effects such as a high variance between the points' axial camera distances may yield different per-point scalers, $\beta_n$. In such circumstances, Eq.~(7) becomes: 
\begin{equation} 
    \mathbf{v}(\mathbf{x}_n,t) = \beta_n \nabla_{\mathbf{x}} u(\mathbf{x}_n, t).
    \label{eq:shifts}
\end{equation}
Thus, substituting the modal expansion (Eq.~(6)) into Eq.~\eqref{eq:shifts} now yields:
\begin{align}
    \mathbf{v}(\mathbf{x}_n,t) &= \gamma\beta_n
                        \sum_{k=1}^{K} \nabla\phi_k(\mathbf{x}_n)\, \big(s(t) * g_k(t)\big) + \eta(\mathbf{x}_n,t) \nonumber \\
                        &= \gamma
                        \sum_{k=1}^{K} \nabla\phi^{\text{aug}}_k(\mathbf{x}_n)\, \big(s(t) * g_k(t)\big) + \eta(\mathbf{x}_n,t),
    \label{eq:main_relation}
\end{align}
where we define the \textit{augmented} mode shape gradient as $\nabla\phi^{\text{aug}}_k(\mathbf{x}_n) := \beta_n \nabla\phi_k(\mathbf{x}_n)$.

Crucially, our procedure does not require explicitly decoupling the varying factors $\beta_n$ from the true physical mode shape gradients $\nabla\phi_k(\mathbf{x}_n)$. Because our mode shape extraction is completely data-driven, it inherently absorbs this per-point scaling factor. Specifically, our numerical recovery of the mode shape gradients is extracted directly using Eq.~\eqref{eq:mode_shapes_recover} (reproduced below):
\begin{equation}
    \nabla\hat\phi_k(\mathbf{x}_n) =
    \operatorname{Re}\left\{
    \frac{\mathbf{V}(\mathbf{x}_n,\hat\omega_k) \cdot
    \mathbf{V}_1(\mathbf{x}_0,\hat\omega_k)^{*}}
    {\mathbb{E}_{n,a}[|\mathbf{V}(\mathbf{x}_n,\hat\omega_k)|]\cdot
    \left|\mathbf{V}_1(\mathbf{x}_0,\hat\omega_k)\right|}
    \right\}.
    \label{eq:mode_shapes_recover}
    \tag{11}
\end{equation}
\vspace{0mm}

Thus, in the case of non-negligible $\beta_n$, Eq.~\eqref{eq:mode_shapes_recover} directly approximates the augmented term $\nabla\phi^{\text{aug}}_k(\mathbf{x}_n)$ up to a global normalization constant. The factors $\beta_n$ will simply skew the numerically recovered mode shape gradients by a constant per-point factor. Consequently, when we invert the model to solve for the recovered latent sound source $\hat{s}^{\text{inv}}(t)$, the optimization (Eq.~\eqref{eq:invS}) remains completely agnostic to the true physical mode shape gradients of the surface:

{\small
\begin{equation}
    \underset{s(t),\alpha_k}{\operatorname{argmin}}\!\!
    \left\|
    \mathbf{v}(\mathbf{x}_n,t) - \sum_{k=1}^{K} \nabla\hat\phi^{\rm aug}_k(\mathbf{x}_n)\, \big(s(t) * \hat g_k(t)\big)
    \right\|_2^2
    \!+\! 
    \lambda\left\| \dot{s}(t) \right\|_2^2.
    \label{eq:invS}
    \tag{12}
\end{equation}
}
\vspace{0mm}

Note that because the latent sound $s(t)$ and the modal impulse responses $\hat{g}_k(t)$ are convolved in our forward model, there is an inherent scale ambiguity. Any remaining global scaling discrepancies between the measured and physical domains are cleanly absorbed jointly by the optimized modal coupling coefficients $\alpha_k$ (implicit within $\hat{g}_k(t)$) and the overall amplitude of the recovered signal $\hat{s}^{\text{inv}}(t)$. Since $s(t)$ is a dimensionless audio signal, this global scale factor simply alters the absolute playback volume without affecting the spectral content or acoustic fidelity. Thus, the precise physical calibration of $\beta_n$ is bypassed without compromising the robust recovery of the acoustic signal.

\figFullResults
\tabViSQOL
\tabSIMRSTFT

\section{Full experimental results}
We extensively evaluated our method on a variety of objects with diverse materials (wood, metal, plastic, rubber), geometries (planar, curved), and irregular shapes (such as the wooden binder and guitar body). 
The approach even works for solid objects like the Yoga block despite not being optimized for volumetric structures.
Figure~\ref{fig:all_results_supp} shows all the objects along with the reconstructed spectrograms. 

\paragraph{Quantitative evaluation}
We further evaluated the recovered audio using several established numerical scores.
Note that we do not have a ground-truth target audio, only the reference source signal played by a speaker in front of each object. Hence, we quantitatively compare the recovered audio to the input source signal.
We compare to three baselines: (i)~\emph{Single}: the raw vibrations measured at one point on the surface, (ii)~\emph{Avg}: naive averaging of all measured vibrations, and (iii)~\emph{DnS}: delay-and-sum where a single global temporal delay is estimated per point~\cite{Benesty2017}.

Table~\ref{tab:visqol-nsim} reports ViSQOLAudio-NSIM~\cite{hines2015visqolaudio, hines2012speech}, which is the raw perceptual similarity index used inside ViSQOLAudio before mean opinion score (MOS) mapping. Higher ViSQOLAudio-NSIM score indicates better similarity between the recovered audio and the input source signal.
Our recoveries consistently score higher than all baselines.
We further used the mean opinion score (MOS) mapping on top of ViSQOLAudio (Tab.~\ref{tab:visqol-mos}). 

Inspired by~\cite{yamamoto2020simrstft, steinmetz2020auraloss}, We further used scale-invariant multi-resolution distance computed in the space of short-time Fourier Transform (STFT). Table~\ref{tab:si-mr-stft} shows the basic distance, while Tab.~\ref{tab:si-mr-stft-p} shows a perceptually-weighed distance computed in this space. On average, our results have smaller distance to the reference input signal compared to all other baselines quantitatively showing the superiority of our audio recovery.

%% file: main.bib
@String(CVPR= {IEEE Conf. Comput. Vis. Pattern Recog.})

@String(ICCV= {Int. Conf. Comput. Vis.})

@String(ECCV= {Eur. Conf. Comput. Vis.})

@String(TOG= {ACM Trans. Graph.})

@String(ICASSP=	{IEEE International Conference on Acoustics, Speech, and Signal Processing (ICASSP)})

@String(ICLR = {Int. Conf. Learn. Represent.})

@String(ICCP = {IEEE International Conference on Computational Photography})

@article{liu2005motion,
  title={Motion magnification},
  author={Liu, Ce and Torralba, Antonio and Freeman, William T and Durand, Fr{\'e}do and Adelson, Edward H},
  journal={ACM transactions on graphics (TOG)},
  volume={24},
  number={3},
  pages={519--526},
  year={2005},
  publisher={ACM New York, NY, USA}
}

@article{wu2012eulerian,
  title={Eulerian video magnification for revealing subtle changes in the world},
  author={Wu, Hao-Yu and Rubinstein, Michael and Shih, Eugene and Guttag, John and Durand, Fr{\'e}do and Freeman, William},
  journal={ACM transactions on graphics (TOG)},
  volume={31},
  number={4},
  pages={1--8},
  year={2012},
  publisher={ACM New York, NY, USA}
}

@article{wadhwa2013phase,
  title={Phase-based video motion processing},
  author={Wadhwa, Neal and Rubinstein, Michael and Durand, Fr{\'e}do and Freeman, William T},
  journal={ACM Transactions on Graphics (TOG)},
  volume={32},
  number={4},
  pages={1--10},
  year={2013},
  publisher={ACM New York, NY, USA}
}

@inproceedings{wadhwa2014riesz,
  title={Riesz pyramids for fast phase-based video magnification},
  author={Wadhwa, Neal and Rubinstein, Michael and Durand, Fr{\'e}do and Freeman, William T},
  booktitle=ICCP,
  pages={1--10},
  year={2014},
  organization={IEEE}
}

@inproceedings{zhang2017video,
  title={Video acceleration magnification},
  author={Zhang, Yichao and Pintea, Silvia L and Van Gemert, Jan C},
  booktitle={Proceedings of the IEEE Conference on Computer Vision and Pattern Recognition},
  pages={529--537},
  year={2017}
}

@inproceedings{elgharib2015video,
  title={Video magnification in presence of large motions},
  author={Elgharib, Mohamed and Hefeeda, Mohamed and Durand, Fredo and Freeman, William T},
  booktitle={Proceedings of the IEEE Conference on Computer Vision and Pattern Recognition},
  pages={4119--4127},
  year={2015}
}

@inproceedings{oh2018learning,
  title={Learning-based video motion magnification},
  author={Oh, Tae-Hyun and Jaroensri, Ronnachai and Kim, Changil and Elgharib, Mohamed and Durand, Fr'edo and Freeman, William T and Matusik, Wojciech},
  booktitle={Proceedings of the European Conference on Computer Vision (ECCV)},
  pages={633--648},
  year={2018}
}

@inproceedings{bouman2013estimating,
  title={Estimating the material properties of fabric from video},
  author={Bouman, Katherine L and Xiao, Bei and Battaglia, Peter and Freeman, William T},
  booktitle={Proceedings of the IEEE international conference on computer vision},
  pages={1984--1991},
  year={2013}
}

@inproceedings{chen2024event,
  title={Event-based motion magnification},
  author={Chen, Yutian and Guo, Shi and Yu, Fangzheng and Zhang, Feng and Gu, Jinwei and Xue, Tianfan},
  booktitle={European Conference on Computer Vision},
  pages={428--444},
  year={2024},
  organization={Springer}
}

@article{davis2014visual,
  title={The visual microphone: Passive recovery of sound from video},
  author={Davis, Abe and Rubinstein, Michael and Wadhwa, Neal and Mysore, Gautham J and Durand, Fredo and Freeman, William T},
  year={2014},
  publisher={Association for Computing Machinery (ACM)},
  journal=TOG,
}

@article{zalevsky2009simultaneous,
  title={Simultaneous remote extraction of multiple speech sources and heart beats from secondary speckles pattern},
  author={Zalevsky, Zeev and Beiderman, Yevgeny and Margalit, Israel and Gingold, Shimshon and Teicher, Mina and Mico, Vicente and Garcia, Javier},
  journal={Optics express},
  volume={17},
  number={24},
  pages={21566--21580},
  year={2009},
  publisher={Optical Society of America}
}

@article{bianchi2019long,
  title={Long-range detection of acoustic vibrations by speckle tracking},
  author={Bianchi, S and Giacomozzi, E},
  journal={Applied optics},
  volume={58},
  number={28},
  pages={7805--7809},
  year={2019},
  publisher={Optical Society of America}
}

@inproceedings{wu2020fast,
  title={Fast motion estimation of one-dimensional laser speckle image and its application on real-time audio signal acquisition},
  author={Wu, Nan and Haruyama, Shinichiro},
  booktitle={2020 the 6th International Conference on Communication and Information Processing},
  pages={128--134},
  year={2020}
}

@article{wu202120k,
  title={The 20k Samples-Per-Second Real Time Detection of Acoustic Vibration Based on Displacement Estimation of One-Dimensional Laser Speckle Images},
  author={Wu, Nan and Haruyama, Shinichiro},
  journal={Sensors},
  volume={21},
  number={9},
  pages={2938},
  year={2021},
  publisher={Multidisciplinary Digital Publishing Institute}
}

@inproceedings{Sheinin:2022:Vibration,
  title={Dual-Shutter Optical Vibration Sensing},
  author={Sheinin, Mark and Chan, Dorian and O'Toole, Matthew and Narasimhan, Srinivasa G.},
  booktitle=CVPR,
  year={2022},
}

@inproceedings{Kichler:2025,
    title={Learning to See Inside Opaque Liquid Containers using Speckle Vibrometry},
    author={Kichler, Matan and Bagon, Shai and Sheinin, Mark},
    booktitle=ICCV,
    year={2025}
}

@inproceedings{Zhang:2023,
  title = {Analyzing Physical Impacts using Transient Surface Wave Imaging},
  author={Zhang, Tianyuan and Sheinin, Mark and Chan, Dorian and Rau, Mark and O'Toole, Matthew
  and Narasimhan, Srinivasa G.},
  booktitle=CVPR,
  year={2023},
}

@inproceedings{cai2025event2audio,
  title     = {Event2Audio: Event-Based Optical Vibration Sensing},
  author    = {Mingxuan Cai and Dekel Galor and Amit Pal Singh Kohli and Jacob L. Yates and Laura Waller},
  booktitle = ICCP,
  year      = {2025},
}

@inproceedings{buyukozturk2016smaller,
  title={Smaller than the eye can see: Vibration analysis with video cameras},
  author={Buyukozturk, Oral and Chen, Justin G and Wadhwa, Neal and Davis, Abe and Durand, Fr{\'e}do and Freeman, William T},
  booktitle={World Conference on Non-Destructive Testing 2016},
  year={2016}
}

@article{chen2017video,
  title={Video camera--based vibration measurement for civil infrastructure applications},
  author={Chen, Justin G and Davis, Abe and Wadhwa, Neal and Durand, Fr{\'e}do and Freeman, William T and B{\"u}y{\"u}k{\"o}zt{\"u}rk, Oral},
  journal={Journal of Infrastructure Systems},
  volume={23},
  number={3},
  pages={B4016013},
  year={2017},
  publisher={American Society of Civil Engineers}
}

@article{wadhwa2016eulerian,
  title={Eulerian video magnification and analysis},
  author={Wadhwa, Neal and Wu, Hao-Yu and Davis, Abe and Rubinstein, Michael and Shih, Eugene and Mysore, Gautham J and Chen, Justin G and Buyukozturk, Oral and Guttag, John V and Freeman, William T and others},
  journal={Communications of the ACM},
  volume={60},
  number={1},
  pages={87--95},
  year={2016},
  publisher={ACM New York, NY, USA}
}

@article{davis2015image,
  title={Image-space modal bases for plausible manipulation of objects in video},
  author={Davis, Abe and Chen, Justin G and Durand, Fr{\'e}do},
  journal={ACM Transactions on Graphics (TOG)},
  volume={34},
  number={6},
  pages={1--7},
  year={2015},
  publisher={ACM New York, NY, USA}
}

@phdthesis{zhou2016vibration,
  title={Vibration Extraction Using Rolling Shutter Cameras},
  author={Zhou, Meng},
  year={2016},
  school={Universit{\'e} d'Ottawa/University of Ottawa}
}

@article{nassi2020lamphone,
  title={Lamphone: Real-time passive sound recovery from light bulb vibrations},
  author={Nassi, Ben and Pirutin, Yaron and Shamir, Adi and Elovici, Yuval and Zadov, Boris},
  journal={Cryptology ePrint Archive},
  year={2020}
}

@book{cremer2005,
  title={Structure-Borne Sound: Structural Vibrations and Sound Radiation at Audio Frequencies},
  author={Cremer, Lothar and Heckl, Manfred and Petersson, Bert A. T.},
  year={2005},
  publisher={Springer-Verlag Berlin Heidelberg},
  edition={3rd}
}

@book{ewins2000,
  title={Modal Testing: Theory, Practice and Application},
  author={Ewins, David J.},
  year={2000},
  publisher={Research Studies Press},
  edition={2nd}
}

@book{rao2007continuous,
  title={Vibration of Continuous Systems},
  author={Rao, Singiresu S.},
  year={2007},
  publisher={John Wiley \& Sons}
}

@book{timoshenko1959,
  title={Theory of Plates and Shells},
  author={Timoshenko, Stephen and Woinowsky-Krieger, S.},
  year={1959},
  publisher={McGraw-Hill},
  edition={2nd}
}

@article{kac1966can_you_hear_the_shape_of_a_drum,
  title={Can one hear the shape of a drum?},
  author={Kac, Mark},
  journal={The american mathematical monthly},
  year={1966}
}

@article{find_peaks_2020scipy,
  title={{SciPy 1.0}: fundamental algorithms for scientific computing in Python},
  author={Virtanen, Pauli and Gommers, Ralf and Oliphant, Travis E and Haberland, Matt and Reddy, Tyler and Cournapeau, David and Burovski, Evgeni and Peterson, Pearu and Weckesser, Warren and Bright, Jonathan and others},
  journal={Nature methods},
  year={2020}
}

@book{meirovitch2001fundamentals,
  title={Fundamentals of Vibrations},
  author={Meirovitch, L.},
  year={2001},
  publisher={McGraw-Hill}
}

@article{savitzky1964smoothing,
  title={Smoothing and differentiation of data by simplified least squares procedures.},
  author={Savitzky, Abraham and Golay, Marcel JE},
  journal={Analytical chemistry},
  year={1964},
  publisher={ACS Publications}
}

@inproceedings{kingma2017adam,
    author={Diederik P. Kingma and Jimmy Ba},
    title={Adam: A Method for Stochastic Optimization},
    booktitle={ICLR},
    year={2015} 
}

@misc{HOLOEYE_DE-R351,
  author       = {{HOLOEYE Photonics AG}},
  title        = {{DE-R 351 Diffractive Optical Element}},
  howpublished = {\url{https://holoeye.com/product/de-r-351/}},
  year         = {2023},
  note         = {Accessed: 2025-11-11}
}

@article{chen201558,
    title = {Modal identification of simple structures with high-speed video using motion magnification},
    journal = {Journal of Sound and Vibration},
    volume = {345},
    pages = {58-71},
    year = {2015},
    author = {Justin G. Chen and Neal Wadhwa and Young-Jin Cha and Frédo Durand and William T. Freeman and Oral Buyukozturk}
}

@InProceedings{Jo_2015_ICCV,
author = {Jo, Kensei and Gupta, Mohit and Nayar, Shree K.},
title = {SpeDo: 6 DOF Ego-Motion Sensor Using Speckle Defocus Imaging},
booktitle = {Proceedings of the IEEE International Conference on Computer Vision (ICCV)},
month = {December},
year = {2015}
}

@article{Smith2017,
author = {Smith, Brandon M. and Desai, Pratham and Agarwal, Vishal and Gupta, Mohit},
title = {CoLux: multi-object 3D micro-motion analysis using speckle imaging},
year = {2017},
issue_date = {August 2017},
publisher = {Association for Computing Machinery},
address = {New York, NY, USA},
volume = {36},
number = {4},
issn = {0730-0301},
url = {https://doi.org/10.1145/3072959.3073607},
doi = {10.1145/3072959.3073607},
journal = {ACM Trans. Graph.},
month = jul,
articleno = {34},
numpages = {12}
}

@InProceedings{Smith_2018_CVPR,
author = {Smith, Brandon M. and O'Toole, Matthew and Gupta, Mohit},
title = {Tracking Multiple Objects Outside the Line of Sight Using Speckle Imaging},
booktitle = {Proceedings of the IEEE Conference on Computer Vision and Pattern Recognition (CVPR)},
month = {June},
year = {2018}
}

@book{Benesty2008,
  author    = {Benesty, Jacob and Chen, Jingdong and Huang, Yiteng},
  title     = {Microphone Array Signal Processing},
  publisher = {Springer Berlin, Heidelberg},
  year      = {2008},
  doi       = {10.1007/978-3-540-78612-2},
  isbn      = {978-3-540-78611-5},
  ean       = {978-3-540-78612-2}
}

@book{Benesty2017,
  author    = {Benesty, Jacob and Cohen, Israel and Chen, Jingdong},
  title     = {Fundamentals of Signal Enhancement and Array Signal Processing},
  publisher = {John Wiley \& Sons Singapore Pte. Ltd.},
  year      = {2017},
  doi       = {10.1002/9781119293132},
  isbn      = {9781119293132}
}

@book{Haykin1984,
  editor    = {Haykin, Simon},
  title     = {Array Signal Processing},
  publisher = {Prentice Hall},
  year      = {1984},
  isbn      = {9780130464828},
  series    = {Prentice-hall Signal Processing Series}
}

@book{Sullivan2008,
  author    = {Sullivan, Mark C.},
  title     = {Practical Array Processing},
  publisher = {McGraw Hill},
  year      = {2008},
  isbn      = {9780071548984},
  pages     = {256}
}

@book{Richards2014,
  author    = {Richards, Mark A.},
  title     = {Fundamentals of Radar Signal Processing},
  edition   = {2nd},
  publisher = {McGraw-Hill Education},
  year      = {2014},
  isbn      = {9780071798337}
}

@techreport{Borg2025,
  author      = {Borg, Jack Denman and Dolan, Dan},
  title       = {Sound Speeds of Solids from Ultrasonic Pulse Receiver Measurements},
  institution = {Sandia National Laboratories},
  year        = {2025},
  doi         = {10.13140/RG.2.2.18031.50080},
  url         = {https://rgdoi.net/10.13140/RG.2.2.18031.50080}
}

@article{Wu:25,
author = {Chaoneng Wu and Wei Li and Yizhi Liang and Peiqian He and Changze Song and Xue Bai and Linghao Cheng and Long Jin and Bai-Ou Guan},
journal = {Biomed. Opt. Express},
number = {5},
pages = {1909--1924},
publisher = {Optica Publishing Group},
title = {Phase-coherent multi-sensor synthesis for enhanced photoacoustic imaging: a comprehensive framework for optimal sensor integration},
volume = {16},
month = {May},
year = {2025},
url = {https://opg.optica.org/boe/abstract.cfm?URI=boe-16-5-1909},
doi = {10.1364/BOE.560286}
}

@INPROCEEDINGS{Howard2023,
  author={Howard, Matthew and Hirakawa, Keigo},
  booktitle={ICASSP 2023 - 2023 IEEE International Conference on Acoustics, Speech and Signal Processing (ICASSP)}, 
  title={Event-Based Visual Microphone}, 
  year={2023},
  volume={},
  number={},
  pages={1-5},
  doi={10.1109/ICASSP49357.2023.10094677}}

@InProceedings{Niwa_2023_CVPR,
    author    = {Niwa, Ryogo and Fushimi, Tatsuki and Yamamoto, Kenta and Ochiai, Yoichi},
    title     = {Live Demonstration: Event-Based Visual Microphone},
    booktitle = {Proceedings of the IEEE/CVF Conference on Computer Vision and Pattern Recognition (CVPR) Workshops},
    month     = {June},
    year      = {2023},
    pages     = {4054-4055}
}

@incollection{HE2001159,
title = {Modal analysis methods – frequency domain},
editor = {Jimin He and Zhi-Fang Fu},
booktitle = {Modal Analysis},
publisher = {Butterworth-Heinemann},
address = {Oxford},
chapter = {8},
pages = {159-179},
year = {2001},
isbn = {978-0-7506-5079-3},
doi = {https://doi.org/10.1016/B978-075065079-3/50008-5},
url = {https://www.sciencedirect.com/science/article/pii/B9780750650793500085},
author = {Jimin He and Zhi-Fang Fu}
}

@book{Cohen2010,
  editor    = {Cohen, Israel and Benesty, Jacob and Gannot, Sharon},
  title     = {Speech Processing in Modern Communication},
  subtitle  = {Challenges and Perspectives},
  year      = {2010},
  publisher = {Springer Berlin, Heidelberg},
  doi       = {10.1007/978-3-642-11130-3},
  isbn      = {978-3-642-11129-7},
  series    = {Springer Topics in Signal Processing}
}

@article{alterman2021imaging,
  title={Imaging with local speckle intensity correlations: theory and practice},
  author={Alterman, Marina and Bar, Chen and Gkioulekas, Ioannis and Levin, Anat},
  journal={ACM Transactions on Graphics (TOG)},
  volume={40},
  number={3},
  pages={1--22},
  year={2021},
  publisher={ACM New York, NY}
}

@inproceedings{steinmetz2020auraloss,
    title={auraloss: {A}udio focused loss functions in {PyTorch}},
    author={Steinmetz, Christian J. and Reiss, Joshua D.},
    booktitle={Digital Music Research Network One-day Workshop (DMRN+15)},
    year={2020}
}

@inproceedings{yamamoto2020simrstft,
  title={Parallel {WaveGAN}: A fast waveform generation model based on generative adversarial networks with multi-resolution spectrogram},
  author={Yamamoto, Ryuichi and Song, Eunwoo and Kim, Jae-Min},
  booktitle=ICASSP,
  year={2020}
}

@InProceedings{Zhou_2025_ICCV,
    author    = {Zhou, Xinyu and Duan, Peiqi and Xiaokaiti, Yeliduosi and Xu, Chao and Shi, Boxin},
    title     = {Event-based Visual Vibrometry},
    booktitle = {Proceedings of the IEEE/CVF International Conference on Computer Vision (ICCV)},
    month     = {October},
    year      = {2025},
    pages     = {24666-24676}
}

@inbook{brinker_carlos_2015_modal_ana,
    author = {Rune Brincker and Carlos E. Ventura},
    publisher = {John Wiley \& Sons, Ltd},
    isbn = {9781118535141},
    title = {Frequency-Domain Identification},
    booktitle = {Introduction to Operational Modal Analysis},
    chapter = {10},
    pages = {261-280},
    doi = {https://doi.org/10.1002/9781118535141.ch10},
    url = {https://onlinelibrary.wiley.com/doi/abs/10.1002/9781118535141.ch10},
    eprint = {https://onlinelibrary.wiley.com/doi/pdf/10.1002/9781118535141.ch10},
    year = {2015},
}

@inproceedings{brincker2000modal,
  title={Modal identification from ambient responses using frequency domain decomposition},
  author={Brincker, Rune and Zhang, Lingmi and Andersen, Palle},
  booktitle={IMAC 18: Proceedings of the International Modal Analysis Conference (IMAC)},
  year={2000}
}

@article{hines2015visqolaudio,
  title={{ViSQOLAudio}: An objective audio quality metric for low bitrate codecs},
  author={Hines, Andrew and Gillen, Eoin and Kelly, Damien and Skoglund, Jan and Kokaram, Anil and Harte, Naomi},
  journal={The Journal of the Acoustical Society of America},
  volume={137},
  number={6},
  pages={EL449--EL455},
  year={2015},
  publisher={AIP Publishing}
}

@article{hines2012speech,
  title={Speech intelligibility prediction using a neurogram similarity index measure},
  author={Hines, Andrew and Harte, Naomi},
  journal={Speech Communication},
  volume={54},
  number={2},
  pages={306--320},
  year={2012},
  publisher={Elsevier}
}


%% file: reference.bib
@String(ICASSP=	{ICASSP})

@String(CVPR  = {CVPR})

@String(ICCV  = {ICCV})

@String(ECCV  = {ECCV})

@String(TOG   = {ACM TOG})

@String(ICLR  = {ICLR})
